\documentclass[10pt,journal]{IEEEtran}
\usepackage{amsmath,amscd,amssymb,amsgen,amsfonts,amsbsy}
\usepackage{mathrsfs}
\usepackage[utf8x]{inputenc}
\usepackage{color}
\usepackage{textcomp}
\usepackage{float}
\usepackage{latexsym,graphicx}

\usepackage{tikz}
\usetikzlibrary{automata,arrows,positioning,calc}
\usepackage{url}
\usepackage{cite}
\usepackage{algorithmic}
\usepackage[ruled,lined]{algorithm2e}
\usepackage{breqn}

\newcommand{\prob}[1]{\mathsf{Pr}\left( #1 \right)}

\newcommand{\remove}[1]{}

\newcommand{\comments}[1]{}

\newcommand{\qed}{\hfill $\square$}

\newcommand{\bm}[1]{\boldsymbol{#1}}
\newtheorem{lemma}{Lemma}
\newtheorem{proposition}{Proposition}

\newtheorem{remark}{Remark}
\newtheorem{definition}{Definition}

\usetikzlibrary{shapes.geometric}

\tikzset{
	buffer/.style={
		draw,
		shape border rotate=120,
		isosceles triangle,
		isosceles triangle apex angle=69,
		fill=white,
		node distance=10cm,
		minimum height=10em
	}
}

\makeatother

\title{Lagrangian Relaxation for  Multi-Action  Partially Observable Restless Bandits: Heuristic Policies and Indexability}	
\author{Rahul Meshram and Kesav Kaza
\thanks{R. Meshram is with the Department of Electrical Engineering, Indian Institute of Technology Madras, Chennai, India. (e-mail: {rahulmeshram}@ee.iitm.ac.in). 
K. Kaza is with the Department of Electrical Engineering and Computer Science, University of Ottawa, Canada (e-mail: kkaza@uottawa.ca)}
}
\begin{document}

\maketitle

\begin{abstract}

Partially observable restless multi-armed bandits have found numerous applications including in recommendation systems, communication systems, public healthcare outreach systems, and in operations research. 
We study multi-action partially observable restless multi-armed bandits, it is a generalization of the classical restless multi-armed bandit problem---1) each bandit has finite states, and the current state is not observable, 2) each bandit has finite actions. In particular, we assume that more than two actions are available for each bandit. We motivate our problem with the application of public-health intervention planning. We describe the model and formulate a long term discounted optimization problem, where the state of each bandit evolves according to a Markov process, and this evolution is action dependent. The state of a bandit is not observable but one of finitely many feedback signals are observable. Each bandit yields a reward, based on the action taken on that bandit. The agent is assumed to have a budget constraint. The bandits are assumed to be independent. However, they are weakly coupled at the agent through the budget constraint.  

We first analyze the Lagrangian bound method for our partially observable restless bandits.
The computation of optimal value functions for finite-state, finite-action POMDPs is non-trivial. Hence, the computation of Lagrangian bounds is also challenging. We describe approximations for the  computation of Lagrangian bounds using point based value iteration (PBVI) and online rollout policy. We further present various properties of the value functions and provide theoretical insights on PBVI and online rollout policy. We study heuristic policies for multi-actions PORMAB.  
Finally, we discuss present Whittle index policies and their limitations in our model. 
%\rahul{rewrite abstract.}
\end{abstract}

\section{Introduction}

\subsection{Motivation} 
Resource allocation under uncertainty is a common problem faced in applications with dynamic environments. Restless multi-armed bandits are sequential decision models that have been studied and applied to resource allocation in various domains such as wireless networks \cite{Ahmad09, LiuZhao10}, wildfire management \cite{kaza2024constrained}, etc. 
In this paper, we focus on multi-action finite state partially observable restless multi-armed bandits also with potential applications to health care resource allocation and planning, among other things. Let us look at the following healthcare planning scenario as a motivating example. Consider a finite state representation describing the health of an individual. The states can be ordered, where the highest state is interpreted as very healthy, and the lowest state as very unhealthy. Often, health care workers can not observe the exact state of an individual, but, can observe signals/symptoms which are dependent on the actions of the healthcare worker such as the questions they ask, the set of tests they administer or the medicines they prescribe. This situation can be represented as a multi-action finite state POMDP model. Individual behavior (with respect to medicine administration/adhering to the prescribed protocol) and health changes over time, and these changes depend on the various interventions of the health care worker. These interventions are constrained due to limited availability of health care workers and medical resources. This motivates the formulation of a resource allocation problem in which a planner must schedule workers $K$ out of $N$ in each round with different actions for health workers given the budget constrained. We model public health interventions using multi-action partially observable RMAB. 

Our model is a generalization of the two action two state partially observable RMAB. It is an important class of problems with applications in many domains such as machine maintenance, online recommendation systems, wireless, opportunistic  communication systems. 
Recently, RMAB has been applied for public health intervention planning \cite{Mate2021banditField}. 
This application is motivated from the observation that intelligent scheduling of health care interventions improves the adherence of patients to medications for diseases like diabetes, hypertension, tuberculosis, HIV, cancer. The essential goal is to keep the health of patient in good state through prevention/early diagnosis and maintaining the adherence to prescribed protocol. 

Finite state representation allows us to capture different levels of severity of  health/ill health, which is not possible using a two-state model. More than two actions in the model describe different levels of intervention from health workers. Moreover, health status is not completely observable, different levels of interventions can provide better information about the health status. Thus, our model considers a finite set of observation signals. The ``higher'' interventions can lead to higher likelihood of observing higher signals, which can reveal more accurate information about the state, i.e., perfect information about the status of health. 
In each round, the reward is a function of the state and the intervention level that is chosen by health workers. The objective of the planner is to schedule health workers with different levels of interventions subject to a budget constraint in each round, such that the long term discounted cumulative reward function is maximized.  

Multi-action finite state partially observable multi-armed restless bandits has applications to communication systems---multiple power level transmission with channel condition when channel is not observable, online recommendation systems \cite{meshram2021monte}, machine replacement problem---there can be multiple actions and states are not observable, \cite{Ross71}.

\subsection{Related work}

RMAB is class of sequential decision problems where the planner schedules ``arms'' sequentially. The state of each arm evolves in time and this evolution is action dependent. The planner has a budget constraint which is usually an integer constraint. Thus, solving this problem is challenging. RMAB was first introduced in \cite{Whittle88}, where Whittle proposed an index based policy that later came to be referred to as the Whittle index policy. Here, indexability is important condition that needs to  be satisfied for each arm, for the application of this policy. This is the simplest form of RMAB, where each arm has two actions--- play or not to play and state of each arm is observable by the planner. In general, RMAB is known to be a PSPACE hard problem \cite{Papadimitriou99}. Although, Whittle index policy is a heuristic policy, it has been shown be optimal under asymptotic conditions under various settings. 
In \cite{Nino-Mora01}, the author introduced a primal-dual based greedy algorithm and studied partial indexability for restless bandits (observable states) using a linear programming approach.

Generalization of RMAB to complex budget constraints was introduced in \cite{hawkins2003lagrangian}, and referred to as weakly coupled Markov decision problem (WC-MDP). Further study on WC-MDP was done by \cite{adelman2008relaxations}. In their model, the state is observable, the actions are finite. They proposed a Lagrangian relaxation approach to the constrained problem by decoupling the WC-MDPs into separate MDPs. 

Partially observable RMAB (PO-RMAB) model has been studied for various applications such as opportunistic communication systems \cite{LiuZhao10}, recommendations systems \cite{Meshram18}, interventions in public health care \cite{Mate2020Bandit,Mate2021Bandit,Mate2021banditField}. All these models consider two states and two actions (play or not to play) for each arm. The state of arm is not observable, hence it is described using a belief state which is updated using Bayes rule. Also, these works have applied and analyzed the Whittle index based policy. 

In \cite{Meshram2021indexability}, the authors studied multi-state (more than two), two-action partially observable RMAB. The belief state is a point in a (probability) simplex, hence, it is difficult to prove indexability and futher difficult to study the Whittle index policy without strong model assumptions. They proposed simulation based heuristic online rollout policy. In \cite{Akbarzadeh2022indexability},  Whittle index policy is studied with strong model assumptions.

We find that there is no study in the literature on multi-state ($>2$) and multi action ($>2$) partially observable RMAB. Classical Whittle index policy is not applicable for this model due to the indexability requirement and the fact that it requires structural results like optimal threshold type policies and complex index computation schemes. It is not possible to compute except under strong model assumptions. Using the Lagrangian relaxation approach, in this paper, we develop a Lagrangian bound on the optimal value function. 

In \cite{kaza2019whittle}, authors studied the Lagrangian bound for two state and two action partially observable RMAB. The computation of the Lagrangian bound is difficult for more than two states and two actions. 

Lagrangian relaxation is a classical method for constrained optimization problems. Using this one can decouple PO-RMAB into finite number of POMDPs. The value function computation of POMDP for finite states and finite actions is difficult, and hence difficulty computation of Lagrangian bound. 

POMDPs have been extensively studied in \cite{Astrom69,Smallwood-Sondik73,Sondik78}. In \cite{Smallwood-Sondik73}, the author introduced the one-pass algorithm based on structural properties of value function which is not feasible for the infinite horizon problem. In \cite{Lovejoy87,Lovejoy91a,Lovejoy91b}, the authors studied properties of value functions and algorithms for POMDP. Point based value iteration (PBVI) which is an approximation to one pass algorithm was developed by \cite{Pineau2006}. This has significantly reduced the complexity of computing the value function. The goodness of approximation depends on number of belief state point selection.

\subsection{Contribution of this paper}

Our contributions are as follows. We formulate the finite state finite action partially observable restless multi-armed bandit (PO-RMAB) problem. Our work is the first to study multi-action PO-RMAB. 
We propose a Lagrangian relaxation technique using Lagrangian multipliers method for budget constraints. We describe the properties of value functions and decouple the problem into $N$ single armed bandits, which are essentially POMDPs. We develop two timescale stochastic approximation based approach for the Lagrangian bound computation. We present PBVI algorithm and its significance for computation of the Lagrangian bound. We also study Monte Carlo online rollout policy for POMDP and its extension to the computation of the Lagrangian bound. We present Lagrangian based heuristic policy and greedy policy. We present a discussion on indexability and the Whittle index policy, and the difficulties in application of these index policies for multi-action PO-RMAB.  

The paper is organized as follows. We present the preliminaries and model description in Section~\ref{sec:model}. We discuss the Lagrangian relaxation approach in \ref{sec:Lagrangian-relaxation}, and approximation to value iteration for POMDP using PBVI in \ref{sec:PBVI}. We next present a study on Monte-Carlo rollout policy for Lagrangian bound in \ref{sec:Monte-carlo-rollout-policy}, heuristic policies in \ref{sec:heuristic-policies}, indexability and Whittle index policy in \ref{sec:indexability}. We finally present a discussion and concluding remarks in \ref{sec:concluding-remark}.

%Structural properties and indexability are developed in Section~\ref{sec:Structural-Prop}. Monte Carlo rollout policy is discussed in Section~\ref{sec:Monte-Carlo-Rollout}, followed by  numerical examples and discussion in Section~\ref{sec:numerical-discussion}.   

%In our work we study the user behavior 

%\subsection{Related Work}

\section{Problem Description}
\label{sec:model}

Consider partially observable restless $N$-armed bandits. The arms are denoted by $n,$ $1 \leq n \leq N$, and are assumed to be independent. The state of each arm is partially observable. Hence, each arm is a partially observable Markov decision process (POMDP), denoted as $\mathcal{M}_{n}= \{\mathcal{S}_n, \mathcal{A}_n, \mathcal{P}_n, \mathcal{R}_n, \mathcal{O}_n, \mathcal{Z}_n, \beta\}$. All arms have $M$ states and $J$ actions. The state space of arm $n$ is $\mathcal{S}_n = \{0,1, \dots, M-1\}$, and the action space is $\mathcal{A}_n = \{0,1,2, \dots, J-1\}$. 
%\textcolor{blue}{
The transition probability matrix $\mathcal{P}_n^{a} = [[p^a_n(i,j)]]$ where $p^a_n(i,j)$ represents the probability of transitioning from state $i$ to state $j$ when action $a$ is taken for arm $n,$ $a \in \mathcal{A}_n.$
%}
Since the state of an arm is not directly observable, the planner maintains a belief about the state, and it is updated based on the observed signals. The planner perceives one among a finite set $\mathcal{O}_n = \{0,1,2,3, \dots, K-1\}$ of $K$ observations.

%\textcolor{blue}{
The probability of observing signal $k \in \mathcal{O}$ from state $i$ under action $a$ for arm $n$ is given by $\rho_{k,n}^{s,a} = \mathbb{P}(o = k~|~s_n=s, A_n= a)$, where $Z_n= [[\rho_{k,n}^{i,a}]]$ represents the observation probability matrix for arm $n$. The observation probabilities are also arm dependent.
%}

The system works in discrete time which is denoted by $t.$
The state of arm $n$ at time $t$ is denoted by $s_{n}(t) \in \mathcal{S}_n$. The planner selects an action for arm $n$ is $a_{n}(t) \in \mathcal{A}_n$ at time $t.$ Then, the reward received from arm $n$ is $r(s_n(t),a_n(t))$ at the time step $t,$ $\mathcal{R}_n$ is denotes the reward matrix for arm $n$. 

%\textcolor{blue}{
Arm $n$ changes it state at each time step $t$ according to the probability $p^{a}_n(s_n,s_n^{\prime}),$ i.e., $\mathbb{P}(s_{n}(t+1) =s_n^{\prime}~|~s_{n}(t) = s_n, a_n(t) =a_n ) = p^{a}_n(s_n,s_n^{\prime}).$ The discount parameter is represented by $\beta.$
%}

An infinite-horizon discounted reward problem with policy $\phi$ is formulated as follows:
\begin{equation}
V_{\phi}(s) = \mathbb{E}_{\phi}\left(\sum_{t=0}^{\infty} \sum_{n=1}^{N} \beta^{t} r_n(s_{n}(t),a_{n}(t)  \right),
\label{eqn:T-horizon-valf-RMAB}
\end{equation}
subject to the budget constraint $\sum_{n=1}^{N} a_n(t) \leq B$ for all $t \geq 0$, where $B$ is the budget. 
%The budget constraint further implies $\sum_{j=0}^{J} z_{i,j,t} =1$ for all $i,t$ and $\sum_{i=1}^{N} \sum_{j=0}^{J} z_{i,j,t} \leq B$ for all $t$.

The policy $\phi$ is defined as a mapping $\phi: H(t)\rightarrow \{a_1,a_2,\dots,a_N\}$, where $H(t)$ denotes the history up to time $t$, given by $H(t):=\{\mathbf{a}(1),\mathbf{o}(1), \dots, \mathbf{a}(t-1),\mathbf{o}(t-1)\},$ $\mathbf{a}(t) = \{a_1(t),\cdots,a_N(t)\},$ and $\mathbf{o}(t) = \{o_1(t),\cdots,o_n(t)\}.$ Since the state is not observable, we define the belief associated for each arm, and it is given as follows.
\begin{eqnarray*}
\omega_{n}^s(t) &=& \prob{s_{n}(t) = s~|~H(t), \bm{\omega}_n(0)}, 
%\\
%&=& \prob{}
\end{eqnarray*}
which represents the probability that arm $n$ is in state $s_n = s$, given past observations, actions, the initial belief vector $\bm{\omega}_n(0),$ and $\bm{\omega}_n(0) = [\omega_n^1(0),\cdots, \omega_{n}^M(0)]^T$
Further, $\sum_{s=0}^{M-1} \omega_{n}^s(t) = 1$ and $\omega_{n}^s(t) \geq 0.$

%\textcolor{blue}{}
The expected reward for arm $n$ is given by 
\begin{eqnarray*}
    R(\boldsymbol{\omega}_n(t),a_{n}(t)) &=& \mathbb{E}[r(s_{n}(t),a_{n}(t))] \\
    &=& \sum_{s \in \mathcal{S}_n} \omega_{n}^s(t) r(s_n(t) = s,a_{n}(t) = a_n)
\end{eqnarray*}
The feasible action set is defined as:
\begin{align*}
    \mathcal{A} = \{ \boldsymbol{a}(t) = (a_{n}(t))_{n=1:N} : a_{n}(t) \in    \{0,1,\dots,J\},
    \\ 
    \sum_{n=1}^{N} a_{n}(t)  \leq B \}.
\end{align*}
The discounted cumulative value function under policy $\phi$ with belief state $\bm{\omega} = (\boldsymbol{\omega}_1,\cdots,\boldsymbol{\omega}_N).$
%Then we have in belief state:
\begin{equation}
V_{\phi}(\bm{\omega}) = \mathbb{E}_{\phi}\left(\sum_{t=0}^{\infty} \sum_{i=1}^{N} \beta^{t} R(\bm{\omega}_n(t),a_{n}(t))  \right),
\label{eqn:T-horizon-valf-RMAB-2}
\end{equation}
The optimal value function is as follows.
\begin{equation*}
V(\bm{\omega}) =  \max_{\phi} V_{\phi}(\bm{\omega})
\end{equation*}
%\textcolor{blue}{
The optimal dynamic program is 
\begin{eqnarray}
    V(\bm{\omega}) = \max_{\bm{a} \in \mathcal{A} } \left[ \sum_{n=1}^{N} R(\bm{\omega}_n,a_{n}) + \beta 
\sum_{\bm{o} \in \mathcal{O}} V(\tau(\bm{\omega},\bm{o},\bm{a}) )\times
\nonumber \right. \\ \left.
\prob{ \bm{o} ~|~\bm{\omega},\bm{a}}
    \right.\Bigg]
\label{Eqn:opt-dynamic-prog}    
\end{eqnarray}
Note that $\bm{\omega}^{\prime}= \tau(\bm{\omega},\bm{o},\bm{a}).$
Since by assumption of independent arms, we have
\begin{eqnarray*}
    \prob{ \bm{o} ~|~\bm{\omega},\bm{a}} = \prod_{n=1}^{N} \prob{o_n~|~\bm{\omega}_n,a_n}
\end{eqnarray*}
Also,
$\bm{\omega}^{\prime} = (\bm{\omega}^{\prime}_1,\cdots,\bm{\omega}^{\prime}_N)$ and $ \bm{\omega}_n^{\prime}= \tau(\bm{\omega}_n,o_n,a_n).$ The computation of belief update is described next.

\subsection{Belief update rule}
\label{sec:belief-update-rule}
We define the history $H(t)$ for all arms and $H(t) = \{ H_1(t), H_2(t),\cdots, H_N(t)\}$ and $H_n(t) = \{a_n(t^{\prime}),o_n(t^{\prime}), \bm{\omega}_n(t^{\prime}) \}_{1 \leq t^{\prime} < t}.$ 
Note that $H_n(t)$ denotes the history of actions, observations, and belief state for arm $n$, $1\leq n \leq N.$ Here, we have assumed that the arms are independent, but, they are weakly coupled through the planner's constraints.

%\textcolor{blue}{
Let $\omega_{n}^s(t) = \prob{ s_{n}(t) = s~|~ H_n(t),a_n(t),o_n(t)}$ be the belief about the state $s$ for arm $n$ at the end of time step $t.$
Note that $\bm{\omega}_n(t-1)$ is a sufficient statistic \cite{BertsekasV195}; hence, we can write 
\begin{eqnarray*}
    \omega_{n}^s(t) &=& \prob{ s_{n}(t) = s~|~ H_n(t),a_n(t),o_n(t)} \\ 
    &=&  \prob{ s_{n}(t) = s~|~  \bm{\omega}_n(t-1),a_n(t),o_n(t)}
\end{eqnarray*}
%Moreover, $\omega_n(t) = \left[\omega^1_n(t),\cdots, \omega^S_n(t) \right]^T $ is the belief vector for arm $n.$
Moreover, $\bm{\omega}_n(t)= (\omega_{n}^1(t), \cdots,\omega_{n}^M(t))^T$ is the belief vector for arm $n.$
%}

Define $\bm{\omega}(t) = [\bm{\omega}_1(t), \cdots, \bm{\omega}_N(t)]$  and it is belief matrix, where each column sums to $1.$ 

We now describe the belief update using Bayes rule. Since the arms are independent, update rule is defined for a single arm. It can be computed for other arms similarly. 

%\subsubsection{Belief update for single arm}

%\textcolor{blue}{
For arm $n$, given that the action for that arm is $a_n(t) = a$, and observation from that arm is $o_n(t) = k$, the previous belief state $\bm{\omega}_n(t),$ the belief update rule for state $s_n(t+1) =s$ at time $t+1$ is given as follows. 
\begin{align*}
   \omega_{n}^s(t+1) = \prob{ s_n(t+1) = s~|~ \bm{\omega}_n(t), a_n(t) = a, o_n(t) = k},
\end{align*}
\begin{align*}
 \omega_{n}^s(t+1)   = \frac{\sum_{s^{\prime}\in\mathcal{S}}\rho^{s^{\prime},a}_{k,n} \omega_{n}^{s^{\prime}}(t)p^a_n(s^{\prime},s)}{\sum_{s^{\prime}\in\mathcal{S}} \omega_{n}^{s^{\prime}}(t)\rho^{s^{\prime},a}_{k,n}}.
\end{align*}
A derivation of this expression using Bayes rule is given in  Appendix A.
%can be found in the appendix.
%}

\section{Lagrangian Relaxation Approach}
\label{sec:Lagrangian-relaxation}
Solving problem~\eqref{Eqn:opt-dynamic-prog} is 
computationally hard. It is a weakly coupled POMDP/PO-RMAB.
It is not separable into independent arms due to constraints. Further, computation of the value iteration algorithm is challenging for partially observable RMAB as the belief space is entire simplex of dimension $M-1$ for states $M.$ 

We develop Lagrangian relaxation approach for weakly coupled PO-RMAB. We further present structural results and two-timescale stochastic approximation based algorithm, where the value function update happens on a faster timescale and the Lagrange multiplier is updated at a slower timescale. 

%\textcolor{blue}{
The Lagrangian relaxation of value function in Eqn.~\eqref{Eqn:opt-dynamic-prog} is introduced by bringing the budget constraint in  the feasible action set $\mathcal{A}$ into the objective function with $\lambda$ as the Lagrangian multiplier for the budget constraint.
{\small{
\begin{eqnarray}
    V^{\lambda}(\bm{\omega}) = \max_{\bm{a} \in \{0,1,2,\cdots,J-1\}^N} \left\{ \sum_{n=1}^{N} R(\bm{\omega}_n,a_n) + \lambda\left(B - \sum_{n=1}^{N}a_n\right)
    \right. \nonumber \\ \left. 
    + \beta \sum_{\bm{o} \in \mathcal{O}} V^{\lambda}(\tau(\bm{\omega},\bm{o},\bm{a})) \prod_{n=1}^{N} \prob{o_n ~|~\bm{\omega}_n,a_n}
    \right\} \nonumber\\ 
    \label{eqn:DP-1}
\end{eqnarray}
}}
\normalsize{}
Here, $\bm{a} = (a_1,\cdots,a_n,\cdots,a_N),$ and $a_n \in \{0,1,\cdots,J-1\}.$

%\textcolor{blue}{
In the preceding equation the optimal value function can be decomposed into value functions of $N$ single-armed bandits, as given by the following Lemma.
\begin{lemma}
We have 
\begin{eqnarray}
   V^{\lambda}(\bm{\omega}) = 
   \sum_{n=1}^{N} V_n^{\lambda}(\bm{\omega}_n) + \frac{B \lambda}{1-\beta}
   \label{eqn:DP-2}
\end{eqnarray}   
where, 
\begin{eqnarray*}
     V_n^{\lambda}(\bm{\omega_n}) = 
    \max_{a_n \in \{0,1,2, \cdots,J-1\}} \left\{
    R(\bm{\omega}_n,a_n) - \lambda a_n 
      \right. \\ \left. 
    + \beta \sum_{o_n \in \mathcal{O}_n}
    V_{n}^{\lambda}(\tau(\bm{\omega}_n,o_n,a_n)) \prob{o_n~| \bm{\omega}_n,a_n}
    \right\}
\end{eqnarray*}
\label{lemma:val-fun-decouple}
\end{lemma}
%}
Proof is given in Appendix~\ref{Proof-lemma:val-fun-decouple-appendix}.
This result is motivated from \cite[Proposition 1]{adelman2008relaxations} which was presented for weakly coupled MDPs. In our model, we extend it for weakly coupled POMDPs.
Here, recursive expansion of Eqn~\eqref{eqn:DP-1} after substitution of RHS of Eqn.~\eqref{eqn:DP-2} can lead to the desired result. It follows from the Bellman optimality equation. 

Further, for any $\lambda \geq 0,$ $V^{\lambda}(\bm{\omega}) \geq V(\bm{\omega})$ for all belief states $\bm{\omega} \in \Delta^{N},$ where $\Delta$ is a simplex of dimension $M-1$ (belief simplex). $\bm{\omega}_n \in \Delta$ for all $n,$ $\bm{\omega} =(\bm{\omega}_1,\cdots,\bm{\omega}_N).$

We define $\mathcal{T}$, the Bellman operator, and $\mathcal{T} V_n(\bm{\omega})$ is given by
\begin{eqnarray}
     \mathcal{T} V_n^{\lambda}(\bm{\omega_n}) := 
    \max_{a_n \in \{0,1,2, \cdots,J-1\}} \left\{
    R(\bm{\omega}_n,a_n) - \lambda a_n 
      \right. \nonumber\\ \left. 
    + \beta \sum_{o_n \in \mathcal{O}_n}
    V_{n}^{\lambda}(\tau(\bm{\omega}_n,o_n,a_n)) \xi(\bm{\omega},o_n,a_n)    \right\} , 
    \label{Eqn:opt-bellman-operator}
\end{eqnarray}
and $\xi(\bm{\omega}_n,o_n,a_n) := 
    \prob{o_n~| \bm{\omega}_n,a_n}.$
%Details of the proof can be found in the Appendix.
Next, we show that the value function is upper bounded by Lagrangian based value function at any given belief state .  
\begin{proposition}
For any $\lambda \geq 0,$ $V^{\lambda}(\bm{\omega}) \geq V(\bm{\omega})$ for all $\bm{\omega} \in \Delta^N.$
\label{Prop:Val-fn}
\end{proposition}
Proof is given in Appendix~\ref{proof:Prop:Val-fn}.  
%\textcolor{blue}{Proof:The idea is to show that the expression in RHS of to substitute Eqn~\eqref{eqn:DP-2} for $V^{\lambda}(\Gamma_{o}(\omega))$ in Eqn.~\eqref{eqn:DP-1} and $V^{\lambda}(\omega).$ Then difference between LHS and RHS of \eqref{eqn:DP-1} has to be $0.$ }

After Lagrangian relaxation, the value function becomes separable for the $N$ armed PO-RMAB. This allows us to compute a Lagrangian bound. In our model, another difficulty is due to partially observable MDPs. In the following, computation of the Lagrangian bound is discussed.

\subsection{Computation of Lagrangian Bound}
The Lagrangian bound is computed by solving the optimization problem with respect to the Lagrangian variable $\lambda.$ We have  
%\textcolor{blue}{
\begin{eqnarray}
    \min_{\lambda \geq 0} V^{\lambda}(\bm{\omega}) = 
   \sum_{n=1}^{N} V_n^{\lambda}(\bm{\omega}_n) + \frac{B \lambda}{1-\beta}
   \label{eqn:Val-lagrange-min}
\end{eqnarray}
%}
This optimization is min-max problem, the minimization is with the dual variable $\lambda \geq 0$ and the maximization is with the  primal variables which are the actions of the bandits using value-iteration.
Thus, it is required to solve the optimal Bellman equation~\eqref{Eqn:opt-bellman-operator}.
This is equivalent to solving for the value function for a POMDP parametrized by $\lambda.$ We now present the properties of the value function.
\begin{lemma}
%\[
\begin{enumerate}
\item $V_n^{\lambda}(\bm{\omega}_n)$ is piecewise-linear and convex  in $\bm{\omega}_n$ for fixed $\lambda.$
\item 
$V_n^{\lambda}(\bm{\omega}_n) $ is piecewise linear, convex, and  decreasing in $\lambda$ for fixed $\bm{\omega}_n.$ Further, as $\lambda \rightarrow \infty$, we have
\begin{eqnarray}
    \frac{\partial V_n^{\lambda}(\bm{\omega}_n)}{\partial \lambda} \rightarrow 0 .
\end{eqnarray}
%\]
\item $V_n^{\lambda}(\bm{\omega}_n)$ is Lipschitz in $\bm{\omega}_n$ with suitable Lipschitz constant. 
\item For $\lambda_l \leq \lambda_{\min} \leq \lambda_u$ we can have
\begin{eqnarray}
    -\sum_{n=1}^{N} \frac{\partial V_n^{\lambda}(\bm{\omega}_n)}{\partial \lambda} \leq \frac{B}{1-\beta}.
\end{eqnarray}
\end{enumerate}
\label{lemma:Vproperties}
\end{lemma}
Proof is by using the principle of induction, and can be found in Appendix D. 

%\textcolor{blue}{ }
%Computation of Lagrangian bound requires the computation of value function $V^{\lambda}_n$ for each arm $n.$ Note that this is parametrized by Lagrangian variable $\lambda.$ 
In Lagrangian bound computation, we employ a two-timescale variant of stochastic approximation algorithms. Here, assuming $\lambda$ as quasi-static parameter, value iteration is performed. Thus value iteration algorithm runs on a ``natural'' timescale. Next, we update the parameter $\lambda$ using finite difference method and this update is performed on slower timescale compare to the value iteration algorithm. Detailed analysis of two timescales algorithm is found in \cite[Chapter $6$]{Borkar08}. 

%\textcolor{blue}{}
%Two timescale algorithms are as follows. 
The value iteration algorithm is given by 
\begin{eqnarray}
   V_{t}^{\lambda_t}(\bm {\omega}) &=& 
   \sum_{n=1}^{N} V_{n,t}^{\lambda_t}(\bm{\omega}_n) + \frac{B \lambda_{t}}{1-\beta},
   \label{eqn:algo-value-iteration1}
\end{eqnarray}
\begin{eqnarray}
    V_{n,t}^{\lambda_t}(\bm{\omega}_n) &=& \mathcal{T}V_{n,t-1}^{\lambda_t}(\bm{\omega}_n).  
   \label{eqn:algo-value-iteration2} 
\end{eqnarray}    

%\textcolor{red}{
Lagrangian multiplier $\lambda_t$ update rule is 
\begin{eqnarray}
\lambda_{t+1} = \left[(1-\eta) \lambda_t + \eta g_t \right]^{+},   
\end{eqnarray}
where 
\begin{eqnarray*}
g_t^{\lambda}(\bm{\omega}) = \frac{\partial V_{t}(\bm{\omega})}{\partial \lambda} = \sum_{n=1}^{N}\frac{\partial V_{n,t} (\bm{\omega}_n)}{\partial \lambda} + \frac{B}{1-\beta},
%\frac{V_{t}-V_{t-1}}{\lambda_t - \lambda_{t-1}}.    
\end{eqnarray*}
%}

Here, $\eta$ is learning rate for $\lambda_t,$  $0<\eta<1$  and it is small. $\left[c\right]^{+} = \max\{c,0\}.$ 

Computation of $\frac{\partial V_{n,t} (\omega_n)}{\partial \lambda}$ is not easy. We compute it using finite difference method. 

%}
%\textcolor{blue}{There is some issue because value function depends on $\omega$ and it implies that $g_t$ depends on $\omega.$}

In the analysis of the two-timescale algorithm, we assume that $\lambda_t = \tilde{\lambda}$ to be constant and analyze the value iteration algorithm.
The value iteration algorithm for POMDP is known to converge to the optimal value function using contraction mapping theorem, and showing that Bellman operator $\mathcal{T}$ is a contraction, \cite{Lovejoy91b}.
Further, the optimal value function is parametrized by  $\tilde{\lambda}.$ 
Hence, 
$||\mathcal{T}V_{n,t}^{\tilde{\lambda}}(\omega) - V_{n}^{\tilde{\lambda}} (\omega)|| \rightarrow 0$ uniformly as $t \rightarrow \infty.$  
For small learning rate $\eta,$ $\lambda$ is quasi-static. 

Now,  $\lambda_{t+1}$ is update is analyzed using stochastic approximations. The limiting ordinary differential equation (ODE) for the $\lambda$ update rule is 
\begin{eqnarray*}
    \dot{\lambda}(t) =  \left( - \lambda(t) + \sum_{n=1}^{N} \frac{\partial V^{\lambda(t)}_n(\omega_n)}{\partial \lambda} + \frac{B}{1-\beta} \right).
\end{eqnarray*}
Note that $V^{\lambda(t)}_n$ converges to the optimal value function and has a unique solution due to the contraction mapping property. Further, $\lambda(t)$ has a unique stable equilibrium and the limiting ODE trajectory converges to the limit set. 
Thus iterate $\lambda_t$ converges to small neighborhood of this equilibrium. The analysis of the two-timescale algorithm is given in \cite[Chapter $6$]{Borkar08}. 
\begin{remark}
\begin{enumerate}
    \item Gradient $g_t$ is difficult to compute, as there is no explicit closed form expression for value function. Hence, we approximate the first term in $g_t$ by a finite difference term. We have provided a two-timescale scheme for Lagrangian bound computation in Algorithm~\ref{algo:Lb-computation}.
    \item The value iteration for POMDP  is difficult to solve as belief state is a point in a probability simplex. However, using properties of value functions, we present a point based value iteration algorithm. 
    %We also study approximate value iteration using Monte Carlo rollout policy.
\end{enumerate}
\end{remark}

%\rahul{Story on two timescale algorithm with update rule.}
%\rahul{Change notations and modify algorithms}
\begin{algorithm}
\caption{Lagrangian Bound (Lb) for PO-RMAB}
\label{algo:Lb-computation}
\begin{algorithmic}[1]
%\STATE \rahul{need to clean this}
\STATE \textbf{Input} 
%$R_n, \rho_n, P_n$ for $n = 1, \dots, N$;
Belief state $\bm{\omega}$; initial Lagrange multiplier $\lambda_0$; tolerance $\delta$; discount factor $\beta$; step sizes $\eta$.
\STATE \textbf{Output} Lagrangian bound $V^{\lambda^*}(\bm{\omega})$, and  optimal $\lambda^*$.

\STATE Initialize $t = 1$, $\lambda_t = \lambda_0$, $V_0^{\lambda} (\bm{\omega}) = \frac{B}{1-\beta} \min\{R_n,0\}$.
\WHILE{true}
    \FOR{$i = 1$ to $N$}
        \STATE Compute $V^{\lambda}_n(\bm{\omega}_n) $ using \text{PBVI Algo}.
    \ENDFOR
    \STATE Compute $V^{\lambda_t}(\bm{\omega}) \leftarrow \frac{B\lambda_t}{1-\beta} + \sum_{n=1}^{N} V^{\lambda_t}_n$.
    \STATE Compute $g_{\lambda_t} \leftarrow \frac{V^{\lambda_t} - V^{\lambda_{t-1}}}{\lambda_t - \lambda_{t-1}}$.
    
    \IF{$|g_{\lambda_t}| \leq \delta$}
        \STATE $V^{\lambda^*} \leftarrow V^{\lambda_t}$, $\lambda^* \leftarrow \lambda_t$.
        \STATE \textbf{break}.
    \ELSE
        \STATE $V^{\lambda} \leftarrow V^{\lambda_t}$.
        \STATE $\lambda_{t+1} \leftarrow \lambda_t + \eta g_{\lambda_t}$.
        \STATE $t \leftarrow t + 1$.
        \STATE \textbf{continue}.
    \ENDIF
\ENDWHILE
\RETURN $V^{\lambda^*}, \lambda^*$
\end{algorithmic}
\end{algorithm}

\section{Approximations: Point Based Value Iteration (PBVI) for POMDP}
\label{sec:PBVI}
The value iteration for POMDP with finite state and finite actions is computationally challenging.  We use the approximation to value iteration algorithm, i.e., PBVI algorithm. In the following, we first present Sondik's one pass algorithm\cite{Smallwood-Sondik73} and later discuss PBVI algorithm \cite{Pineau2003PBVI}. Both these algorithms are applicable when the value function is piece-wise linear and convex in belief state.
 This is developed for each single-armed bandit, as each bandit is a POMDP. As we will deal with single-armed bandits, the explicit dependence of value function on the arm index $n$ and $\lambda$ is omitted for notational simplicity.
We denote the belief as $\bm{\omega}.$ Here, belief $\bm{\omega}$ is an $M$ dimensional vector instead of a matrix, unlike in the earlier section. 

\subsection{Sondik's  One Pass Algorithm}

We represent the value function as the maximum over inner product of finite set of linear functions and parametrized 
$\alpha$ vector. Here, $\alpha$ is an $M$ dimensional vector, $\alpha \in \Gamma,$ and 
$\Gamma$ is set of $\alpha$ vectors. The value function is given by
\begin{equation}
    V(\omega) = \max_{\alpha \in \Gamma} \langle \alpha, \bm{\omega} \rangle = \max_{\alpha \in \Gamma} \sum_{s \in \mathcal{S}} \alpha(s) \omega^s.
\end{equation}
Here, $\alpha = [\alpha(1), \cdots,\alpha(s)]^T $ and $\bm{\omega} = [\omega^1, \cdots,\omega^M]^T.$ $T$ denotes a transpose of a vector.  
%In Sondik's one pass algorithm, it is assumed that piecewise linear and convex. The value functions over belief space is represented using $\alpha$ vectors. 

Consider for any horizon $t,$ the $\Gamma_t = \{\alpha_1,\alpha_2,\cdots,\alpha_m\}$ is set of $\alpha$ vectors. Then, the value function 
\begin{eqnarray}
    V_t(\omega) = \max_{\alpha \in \Gamma_t}  \langle \alpha, \bm{\omega} \rangle .
\end{eqnarray}
%and let $\Gamma_{t-1}$ is 

From dynamic program of a single-armed bandits in Eqn.~\eqref{Eqn:opt-bellman-operator}, we have the following value iteration scheme 
\begin{eqnarray*}
    V_t(\omega) = \max_{a} \left\{ \widetilde{R}(\bm{\omega},a,\lambda) + \beta \sum_{o} \xi(\bm{\omega},o,a) V_{t-1} (\tau(\omega,a,o))\right\}.
\end{eqnarray*}
%$ V_{n}^{\lambda}(\tau(\bm{\omega}_n,o_n,a_n)) \xi(\bm{\omega},o_n,a_n)$
Here, $\widetilde{R}(\bm{\omega},a,\lambda) = R(\bm{\omega},a) - \lambda a.$
After simplification and using $\alpha$ vector set $\Gamma_{t-1}$ at $t-1,$ we obtain  
\begin{eqnarray*}
    V_t(\bm{\omega}) = \max_{a}\left\{ \widetilde{R}(\bm{\omega},a,\lambda) + \beta 
    \sum_{o} \max_{\alpha \in \Gamma_{t-1}} 
    \right. \\ \left. 
    \sum_{s} \sum_{s^{\prime}} 
    \prob{s^{\prime}~|~s,a} \prob{o~|s^{\prime},a}
    \alpha(s^{\prime}) \omega^s   \right\}.
\end{eqnarray*}
%\textcolor{blue}{
It is difficult to compute $V_t(\bm{\omega})$ for all $\bm{\omega} \in \Delta.$ However, the set $\Gamma_t$ can be generated using the set $\Gamma_{t-1}.$ The steps are described in the following. \\ 
Step $1:$  Generate sets $\Gamma_t^{a,*}$  and $\Gamma_t^{a,o}:$
    \begin{eqnarray*}
     \Gamma_t^{a,*} \leftarrow & \alpha^{a,*}(s) = \overline{R}(s,a,\lambda) \ \ \ \ \ \ \ \ \ \ \ \ \ \ \ \ \ \  \ \ \ \ \ \ \ \ \ \ \  \ \ \ \  \ \ \ \  \\
     \Gamma_t^{a,o} \leftarrow & \alpha_i^{a,o}(s) = \beta \sum_{s^{\prime}} \prob{s^{\prime}~|~s,a}\prob{o~|~s^{\prime},a} \alpha_i(s^{\prime}).
    \end{eqnarray*}
    Here, $ \alpha^{a,*}$ and $\alpha_i^{a,o}$ is $M$-dimensional hyper-plane, $\overline{R}(s,a,\lambda) = r(s,a) - \lambda a.$ \\
    Next step is to generate $\Gamma^a_t$ by cross-sum over observations: 
    \begin{eqnarray*}
        \Gamma_t^a = \Gamma_t^{a,*} + \Gamma_t^{a,0} \oplus \Gamma_t^{a,1}  \oplus \Gamma^{a,2}  \oplus \cdots \oplus \Gamma_t^{a,K-1}. 
    \end{eqnarray*}
    Then 
    \begin{eqnarray}
        \Gamma_t = \cup_{a \in \mathcal{A}}\Gamma_t^a.
    \end{eqnarray}
We compute $ V_t(\bm{\omega}) = \max_{\alpha \in \Gamma_t} \sum_{s} \alpha(s) \omega^s.$ 
The computation complexity of the value function is $O(S^2J|\Gamma_{t-1}|^K).$
%}

\subsection{Point Based Value Iteration}

%We follow this paper \cite{Pineau2006}. 
%\rahul{Need to rewrite the following section}
%Solving Partially Observable Markov Decision Processes (POMDPs) exactly is computationally intractable due to the continuous belief space. 

PBVI is an approximate value iteration scheme and the value function is considered for a finite set of belief points. The idea is to iteratively update the value function only at these sampled beliefs. This leads to a reduction in the computational complexity of the value function.
We follow PBVI algorithm from \cite{Pineau2006,Pineau2003PBVI}.

%\emph{Point-based value iteration (PBVI)} is an approximate value iteration algorithm and it operates on a finite set of representative belief points. The key idea is to iteratively update the value function only at these sampled beliefs, significantly reducing computational complexity while retaining good performance.

Let $\mathcal{B}$ be the finite set of sampled belief points, say $m$ points, $\mathcal{B} =\{\bm{\omega}_1,\bm{\omega}_2,\cdots, \bm{\omega}_m \},$ $\forall i=1,2,\cdots,m,  \bm{\omega}_i \in \Delta.$ In 
PBVI, the value function is described using a set of $\alpha$-vectors for each belief point. Thus, the point based value functions are represented by $\{\alpha_1,\alpha_2,\cdots,\alpha_m \}.$ It is a linear function over the belief space, $\mathcal{B}.$ At each iteration, the algorithm performs a backup operation at each belief point in $\mathcal{B}$. Then, we compute new $\alpha$-vectors based on Bellman updates and selecting the one that maximizes expected value.
Steps involved in PBVI algorithm are as follows:

 \begin{enumerate}
     %\item Initialization step: $\Gamma_0 = \{ \alpha_a~|~ \alpha^s_a = R(s,a), \forall a \in \mathcal{A} \}$ and $V_0(\omega) = \max_{a} \sum_{s \in \mathcal{S}} \alpha^s_a \omega^s$ for all $\omega \in \mathcal{B}.$
     \item Step $1:$ We obtain set $\Gamma_t$ (set of $\alpha$ vectors) from the previous set $\Gamma_{t-1},$ and it is done by generating intermediate sets $\Gamma_t^{a,*}$ and $\Gamma_t^{a,o}$ for all $a \in \mathcal{A}$ and $o \in \mathcal{O}.$
     \begin{eqnarray*}
         \Gamma_t^{a,*} \leftarrow & \alpha^{a,*}(s) = \overline{R}(s,a,\lambda) \ \ \ \ \ \ \ \ \ \ \ \ \ \ \ \ \ \  \ \ \ \ \ \ \ \ \ \ \ \ \  \\ 
         \Gamma_t^{a,o} \leftarrow & \alpha_i^{a,o}(s) = \beta \sum_{s^{\prime}} p_{s,s^{\prime}} ^{a} \rho_{s,o}^{a} \alpha_i(s^{\prime}), \forall\alpha_i \in \Gamma_{t-1}.
     \end{eqnarray*}

     \item Next step is to construct $\Gamma_t^{a}$ for all $a \in \mathcal{A}:$
     \begin{eqnarray*}
         \Gamma_t^{a} \leftarrow \alpha_{\omega}^{a} = \Gamma_t^{a,*} + \sum_{o \in \mathcal{O}} \arg\max_{\alpha \in \Gamma_{t}^{a,o}} \left[ \sum_{s} \alpha (s) \omega^s\right],\\ \forall \bm{\omega} \in \mathcal{B},
     \end{eqnarray*}
     $\bm{\omega} = \{\omega^,\cdots,\omega^s,\cdots,\omega^M \}$ and $\bm{\omega} \in \Delta.$
     \item Step $3:$ Find the best action for $\omega \in \mathcal{B}:$
     \begin{eqnarray*}
         \alpha_{\omega} = \arg\max_{\alpha \in \Gamma_t^{a}, \forall a \in \mathcal{A}} \left[\sum_{s}\alpha(s) \omega^s\right]
     \end{eqnarray*}
     and $\Gamma_t = \cup_{\omega \in \mathcal{B}}\alpha_{\omega}.$ Then the value function $V_t(\bm{\omega}) = \max_{\alpha \in \Gamma_t}\sum_{s} \alpha(s)\omega^s.$
     \item The computational complexity of updating value function of set of points $\mathcal{B}$ is polynomial of $|S||A||\Gamma_{t-1}||\mathcal{B}|.$
     %\item $\alpha_a^{s,o} = $
    %\item Suppose $\alpha$ vectors from $\Gamma_t$ and goal is to compute the $\alpha_{t+1}$ and construct the set $\Gamma_{t+1}.$ The recursive step is as follows.
     %\begin{eqnarray}
     %    aa
    % \end{eqnarray}
 \end{enumerate}
%\begin{eqnarray}
 %   \Gamma^{a,*} \leftarrow & \alpha^{a,*} = R(s,a) \\
  %  \Gamma^{a,o} \leftarrow & \alpha^{a,o}_{i} = \beta \sum_{s^{\prime}}T(s,a,s^{\prime}) \Omega(o,s^{\prime},a)\alpha_i^{\prime}(s^{\prime}), \forall \alpha_i^{\prime} \in V^{\prime}
%\end{eqnarray}
%Then one can obtain for all $b \in B,$ for all $a \in A$
%\begin{eqnarray}
%    \Gamma_b^a = \Gamma^{a,*} + \sum_{o} \arg\max_{\alpha \in \Gamma^{a,o}}(\alpha\cdot b).
%\end{eqnarray}
%Finally on can find the best action for each belief point
%\begin{eqnarray}
 %   V \leftarrow \arg\max_{\Gamma_b^a, \forall a \in A}(\Gamma_b^a\cdot b), &\forall b \in B.
%\end{eqnarray}

\textbf{Results}
$R_{\max}:= \max_{s,a} \overline{R}(s,a,\lambda)$ and $R_{\min} : = \min_{s,a} \overline{R}(s,a,\lambda) $
The estimate of value function is denoted $V_t^{\mathcal{B}}$ for belief set $\mathcal{B}$ and horizon $t$ and the optimal value function is denoted by $V^*.$ Then, one want to show that difference $||V_t^{\mathcal{B}} - V^*||_{\infty}$ is bounded. $V_t^{*}$ is the $t-$horizon optimal solution.
Moreover,  difference $||V_t^{\mathcal{B}} - V^*_t||_{\infty}$  goes to zero if $\mathcal{B}$ sample belief increased and densely describe the belief simplex $\Omega.$  
The error in PBVI is given by 
\begin{eqnarray}
    ||V_t^{\mathcal{B}} - V^*||_{\infty} \leq ||V_t^{\mathcal{B}} - V_t^*||_{\infty} + || V_t^* - V^*||_{\infty} 
\end{eqnarray}
Note that $|| V_t^* - V^*||_{\infty}  \leq \beta^{t} || V_0^* - V^*||_{\infty} .$ 
Let $\mathcal{T}$ denotes an exact value backup and $\widetilde{\mathcal{T}}$ denotes the PBVI backup. The error introduced by one iteration of point based backup is 
\begin{eqnarray}
  \epsilon(\bm{\omega})  = || \widetilde{\mathcal{T}} V^B(\bm{\omega}) - \mathcal{T}  V^B(\bm{\omega}) ||_{\infty}
\end{eqnarray}
The maximum total error introduced by point based back is 
\begin{eqnarray}
    \epsilon = \max_{\bm{\omega} \in \Delta} || \widetilde{\mathcal{T}} V^B(\bm{\omega}) - \mathcal{T}  V^B(\bm{\omega}) ||_{\infty} 
\end{eqnarray}
Define the distance $\delta_{\mathcal{B}}$ as follows. 
\begin{eqnarray}
    \delta_{\mathcal{B}} = \max_{\bm{\omega}^{\prime} \in \Delta} \min_{\bm{\omega} \in \mathcal{B}} ||\bm{\omega} - \bm{\omega}^{\prime}||_1
\end{eqnarray}
The error introduced using PBVI during one iteration of value back up over $\mathcal{B}$ is bounded by
\begin{eqnarray}
    \epsilon \leq \frac{(R_{\max}-R_{\min})\delta_{\mathcal{B}}}{(1-\beta)}. 
\end{eqnarray}
Thus for any belief set $\mathcal{B}$ any horizon $t,$ the error of PBVI is given by 
\begin{eqnarray*}
        e_t \leq \frac{(R_{\max}-R_{\min})\delta_{\mathcal{B}}}{(1-\beta)^2}
    \end{eqnarray*}
In \cite{Pineau2006}, various methods for the selection of belief points have been proposed (e.g. set of reachable beliefs, random belief selection).
%\begin{lemma}
%    PBVI error function for per iteration
%    \begin{eqnarray*}
%        \epsilon \leq \frac{(R_{\max}-R_{\min})\delta_B}{(1-\beta)}
%    \end{eqnarray*}
%\end{lemma}

%The error at any time $t,$ $e_t$ is 
    
%Once 

\section{Monte-Carlo Rollout Policy}
\label{sec:Monte-carlo-rollout-policy}
%\rahul{Today need to workout on this section}

We now propose an alternative heuristic rollout policy for the computation of value functions. There can be $N$ rollout policies for $N$ different arms and we compute an approximation of the value function. Note that the approximate value function is dependent on parameter $\lambda$ and it is assumed to be fixed in the rollout policy. This approximate value function is used for Lagrangian bound evaluation in Algorithm~\ref{algo:Lb-computation}.

The Monte Carlo rollout policy  is simulation based approach, and further it is online policy. We obtain approximation to  $V_{n}^{\lambda}(\bm{\omega}_n),$  for $\bm{\omega} \in \Delta$  using rollout policy. For notational simplicity, we omit dependence of value function on the arm subscript $n.$

Given the belief vector $\bm{\omega}$ and the parameter $\lambda,$ the approximate value function is obtained in the following. 
We simulate $L$ trajectories, and a trajectory starts with initial belief vector $\bm{\omega},$ action $a \in \mathcal{A}.$ Each trajectory is simulated for $H$ horizon length.
In each trajectory, we employ a policy $\phi$ and the information collected over a $l$th trajectory is  
\begin{eqnarray*}
\left\{ (\bm{\omega}_{1,l},a_{1,l}, o_{1,l},r_{1,l}), (\bm{\omega}_{2,l}, a_{2,l},o_{2,l}, r_{2,l}), \cdots, 
\right. \\ \left. 
(\bm{\omega}_{H,l}, a_{H,l}, o_{H,l}, r_{H,l}) \right\}
\end{eqnarray*}
The value estimate from $l$th trajectory starting from belief state $\bm{\omega},$ action $a \in \mathcal{A}$ is 
\begin{eqnarray*}
	Q_{H,l}^{\phi,\lambda}(\bm{\omega}, a) = \sum_{h=1}^{H} \beta^{h-1} R_{h,l}^{\phi,\lambda} 
	= \sum_{h=1}^{H} \beta^{h-1} R^{\phi}(\bm{\omega}_{h,l},a_{h,l},\lambda).
	%~|~\omega_{1,k} = \omega, a_{1,k} =a).
\end{eqnarray*}  
Then value estimate over $L$ trajectories is 
\begin{eqnarray*}
	\widetilde{Q}_{H,L}^{\phi,\lambda}(\bm{\omega}, a ) = \frac{1}{L}\sum_{l=1}^{L}  Q_{H,l}^{\phi,\lambda}(\bm{\omega}, a).
\end{eqnarray*} 
The output under policy $\phi$ is $\widetilde{V}_{\phi,H,L}^{\lambda}(\bm{\omega})$  
%
%{\small{
\begin{eqnarray*}
\widetilde{V}_{\phi,H,L}^{\lambda}(\bm{\omega}) = R(\bm{\omega},a,\lambda) + \beta \widetilde{Q}_{H,L}^{\phi,\lambda}(\bm{\omega},a), 
%\\
%\widetilde{V}_{\phi,H,L}(\omega,a= 0,W) =   W + r(\omega,a=0)  +  \widetilde{Q}_{H,L}^{\phi}(\omega, a=1, W) 
\end{eqnarray*}
%}}
Here $a = \phi(\omega) .$
Belief vector update require  $O(S^3)$ computations.
The rollout policy has a worst case complexity $ O(JHL).$ 

Thus, the total computational complexity of rollout policy for $N$ armed hidden Markov restless bandit is $O(NJHL).$ The advantage of rollout is that one can run rollout policies in parallel for $N$ armed bandits. Using the Hoeffding inequality, one can derive conditions on thenumber of trajectories $L$  that are required to measure the goodness of rollout policy for every arm.
\begin{eqnarray}
  \bigg \vert V_{\phi}^{\lambda}(\bm{\omega}) -  \widetilde{V}_{\phi,H,L}^{\lambda}(\bm{\omega}) \bigg \vert \leq \epsilon
\end{eqnarray}
and $L := \frac{2\epsilon^2(1-\beta^2)}{(R_{\max}- R_{\min})^2(1-\beta^H)\log(2/\delta)}.$

One we compute the value function approximation, next step is to improve the policy using policy improvement step.
\begin{eqnarray}
    \widetilde{\phi}(\bm{\omega}) = \arg\max_{a} \left[ R(\bm{\omega},a,\lambda) + \beta \widetilde{Q}_{H,L}^{\phi,\lambda}(\bm{\omega},a) \right]
    \label{eqn:policy-improvement}
\end{eqnarray}
By running the rollout policy and policy improvement step, we can find the better policy than base policy $\phi.$ Computing these for all $\bm{\omega} \in \Delta$ is challenging. One can take finite number of belief point set $\mathcal{B}$  and run rollout policies and this reduces computation.

\section{Heuristic Policies}
\label{sec:heuristic-policies}
%\rahul{Need to write on this section}

In the following, we discuss heuristic policies for solving multi-action PO-RMAB. Solution of the exact problem is intractable. 
We present a Lagrangian based heuristic policy and greedy heuristic policy.

\subsection{Lagrangian Based Heuristic Policy}
%\rahul{Discussion with kesav on Saturday}
We compute the policy for Lagrangian relaxation of the problem. It is computationally challenging because it is an integer programming problem. We solve using a two step approach. Assuming $\lambda$ to be fixed, compute the optimal policy for all arms with the budget constraint. The next step is to find optimal $\lambda.$ The optimal policy is given in the following Lemma. 
\begin{lemma}
Given belief state $\bm{\omega}$ and fixed $\lambda,$ the optimal policy is as follows.
\begin{eqnarray*}
\bm{a}^{*}(\bm{\omega},\lambda) = \arg \max_{\bm{a} \in \mathcal{A} } \left[ \sum_{n=1}^{N} \left(R(\bm{\omega}_n,a_{n}) + \beta 
\right. \right. \\ \left. \left. 
\sum_{o_n \in \mathcal{O}_n}
    V_{n}^{\lambda}(\tau(\bm{\omega}_n,o_n,a_n)) \xi(\bm{\omega},o_n,a_n)
\right) \right],
\end{eqnarray*} 
where $\bm{\omega} = (\bm{\omega}_1, \bm{\omega}_2, \cdots, \bm{\omega}_N),$  $\bm{\omega} \in \Delta^N$ and 
\begin{align*}
    \mathcal{A} = \{ \boldsymbol{a}(t) = (a_{n}(t))_{n=1:N} : a_{n}(t) \in    \{0,1,\dots,J\},
    \\ 
    \sum_{n=1}^{N} a_{n}(t)  \leq B \}.
\end{align*}
\label{lemma:opt-policy}
\end{lemma}
Proof of Lemma is given in  Appendix~\ref{proof:lemma:opt-policy}. 

Observe that $\bm{a}^{*}(\bm{\omega},\lambda)$
is a function of $\lambda.$ Due to coupled constrained in $\mathcal{A}$ it is difficult to compute $\bm{a}^*.$  Further, it has to be optimized with $\lambda.$ Using approach from \cite[Hawkins PhD Thesis 2003, Page No 45]{hawkins2003lagrangian}, we propose the following heuristic algorithm for policy computation.
\begin{enumerate}
    \item Assume that $\lambda$ is fixed and compute the decision for each arm $i,$ $a_n^*(\bm{\omega_n},\lambda)$. PBVI or Rollout policy is used for approximate value function computation:  
    \begin{eqnarray}
        a_n^*(\bm{\omega}_n,\lambda) = \arg\max_{a_n} L_n(\bm{\omega}_n, a_n,\lambda ) 
    \end{eqnarray}
    Here, 
    \begin{eqnarray*}
        L_n(\bm{\omega}_n,a_n,\lambda )  =  R(\bm{\omega}_n,a_{n}) + \beta   \sum_{o_n \in \mathcal{O}_n}
    V_{n}^{\lambda}(\tau(\bm{\omega}_n,o_n,a_n)) \\\times \xi(\bm{\omega},o_n,a_n). 
    \end{eqnarray*}
    This can be solved in parallel for every arm.
    \item Earlier step is repeated for $\lambda_L<\lambda<\lambda_U$ with fixed grid size $\Lambda$.
    \item Hence, we obtain decision vector $\{a_n^*(\bm{\omega}_n,\lambda)\}$ for all arms and for all points on the $\Lambda$ grid.
    \item Find minimum $\lambda$ such that for a given $\bm{\omega},$ $\sum_{n=1}^{N} a_n^*(\bm{\omega}_a,\lambda) \leq B.$ That is, 
    \begin{eqnarray*}
        \min & \lambda \\
        \textbf{s.t} &  \sum_{n=1}^{N} a_n^*(\bm{\omega}_i,\lambda) \leq B \\
        & \lambda \geq 0.
    \end{eqnarray*}
    \item This minimum $\lambda^*$ is the optimal Lagrangian parameter and the optimal decisions are $a_n^*(\bm{\omega}_n,\lambda^*)$
\end{enumerate}
This algorithm's computation time for the optimal policy depends on the underlying value function approximation scheme (PBVI or Rollout) and size of the grid $\Lambda.$

\subsection{Greedy Policy}
We present a simple greedy policy based on immediate reward rather than value function computation. The greedy policy can be combined with online rollout policy and a new look-ahead rollout policy can be studied. Here, we discuss only the greedy policy. 

%\textcolor{blue}{
%Th greedy policy is based on belief state of all arms and reward matrix. 
Let $\bm{\omega}$ be a belief matrix of dimension $M\times N$ and $\mathcal{R}$ be a reward matrix of dimension $M \times J \times N.$ Here, $\bm{\omega} \in \Delta^N.$
The greedy policy selects actions for each armed based on belief $\omega$ and $\mathcal{R}$ at each time step. We have budget constraint $\sum_{i=1}^{N} a_i  \leq B.$ This is a knapsack optimization problem and it is given by 
\begin{eqnarray}
    \max & \sum_{n=1}^{N}  R(\bm{\omega}_n,a_n) \nonumber \\
    \text{s.t.} & 
     \sum_{n=1}^{N} a_n \leq B \nonumber \\
    & a_n \in \{0,1,\cdots, J \} ~\forall~ n =1,2,\cdots,N. 
    \label{opt:greedy}
\end{eqnarray}
This problem is challenging due of the integer constraints. Hence, a greedy heuristic policy is studied.
The immediate expected reward for arm $n$ under action $a_n$ is as follows. $R(\boldsymbol{\omega}_n,a_n) = \sum_{s_n \in \mathcal{S}_n} r_n(s_n,a_n) \omega_{n}^{s_n}.$
%}
%\textcolor{blue}{
\begin{algorithm}
\caption{Greedy Algorithm for Multi-action PO-RMAB}
\begin{algorithmic}[1]
\STATE \textbf{Input:} Belief state matrix $\bm{\omega}$,  reward matrix $\mathcal{R}$ and total maximum budget $B$
\STATE Initial available budget $B_0= B$ and $k=0$
\WHILE{$B_k > 0$ (Budget is positive)} 
    \STATE Compute $R(\boldsymbol{\omega}_n,a_n)$ for $a_n \in \mathcal{A}$ and  $n=1,2,\cdots,N$  
    %Compute the expected immediate reward for each arm and each action of the arm
    \STATE Obtain the action and arm with the highest reward.
    $a_n^* =  \arg\max_{a_n} R(\boldsymbol{\omega}_n,a_n)$ and 
$ n^*=\arg\max_{n} R(\boldsymbol{\omega}_n,a_n)$
    %for each arm based on reward
    \STATE Selected action and arm $\overline{A} = \{(a_{n^*}^{*}, n^*)\}$
    \IF {$a_{n^*}^{*} \leq B_k$ Within budget} 
     \STATE Add action and arm to set $I = I \cup \{(a_{n^*}^{*},n^*) \}$
     \STATE Remove arm $\mathcal{N} = \mathcal{N} - \{i^*\}$
     \STATE Budget reduction $B_{k+1} = B_k -a_{n^*}^{*}$
     \ELSE  
     \STATE Outside budget ($a_{n^*}^{*} > B_k)$
     \STATE Remove Action from playlist $\mathcal{A}^{\prime} = \mathcal{A}^{\prime} - \{a_{n^*}^{*}\}$
    \ENDIF
    %best action of best arm is less than or equal to available budget, then, Include in the set $I$ (arm and action) selected.
    %Next remove that arm from available arms to plays
    %Also Reduce budget by best action $a^{\prime}$
    %
   % \STATE best action of best arm  greater than available budget, then, remove that action from playlist.
    \STATE $k = k+1$ 
\ENDWHILE
\STATE Output Set $I$ (Arms, actions)
\end{algorithmic}
\label{algo:greedy}
\end{algorithm}
%}
%\textcolor{red}{Rahul: minor corrections need in the algorithm. Check whether algorithm is okay or not?} 
%\rahul{This is our one contribution.}
%\textcolor{blue}{
For given belief $\bm{\omega}_n$ find the action with best immediate reward $a_n^* =  \arg\max_{a_n} R(\boldsymbol{\omega}_n,a_n)$
and best immediate reward $R_n^* =  \max_{a_i}R(\boldsymbol{\omega}_n,a_n)$. 
Next, the arm selected is 
$ n^*=\arg\max_{n} R_n^*$ and for this selected arm, the action is $a_{n^*}^*.$
Check with the remaining budget, if $a_{n^*}^* < B_t$. Then include this action in the set $F = \{a_{n^*}^* \}$. The remaining budget is $B_{t+1} = B_t-a_{n^*}^*.$
For remaining arms repeat the procedure. It is described in Algorithm~\ref{algo:greedy}.
%}

The intuition behind this is that we have matrix of dimension $J \times N,$ and entries in this matrix are immediate expected rewards for given belief $\bm{\omega}$ for all arms.
We move along actions for each arm, and find the best action using this reward, and also find the best arm. 
Pick that arm. If this is less than budget available, select in into our box.
Next consider other arms $J \times N-1$
Repeat earlier procedure: 
pick the arm and action, if this action is above the budget.
Reduce matrix dimension to $J-1 \times N-1$. Repeat this procedure until available budget is nil. 
This is a simple greedy procedure which depends on the immediate expected reward.
%We can extend this idea to online rollout policy also.

\section{Indexability and Whittle Index Policy}
\label{sec:indexability}
In this section we discuss about the indexability of PO-RMAB for multi-state and multi action model.  

For two action PORMAB (two states) indexability is well defined.  It is minimum subsidy needed so that not playing an arm becomes equally good as playing, in terms of the value function. This requires computation of the value function. Arms with the highest indices are played. Intuitively, it means that the arms with highest indices can have higher reward in long run. Though it is a heuristic policy, it is shown to be asymptotically optimal or near optimal. Challenges to the use  of this index policy is indexability, which is key requirement. Showing indexability for two-state and two actions PO-RMAB is relatively easy when any one of action provides perfect state information, and it is difficult to claim when any action doesn't provide perfect state information. In special cases it is true, \cite{Meshram18}. It requires structural assumptions on the model. Recently, it is extended for multi-state two action PO-RMAB, the indexability is shown when one of action provides perfect state information. Indexability is proved under structural assumptions on the model. \cite{Meshram2021indexability,Akbarzadeh2022indexability}.

\subsection{Two-action PO-RMAB}
We omit the dependence of arm on index $n,$ and indexability is discussed for a single-armed restless bandit. 
For the sake of clarity, we first discuss two action finite state model and define the index, and conditions for indexability. 

From Lemma~\ref{lemma:val-fun-decouple}, the dynamic program for individual arm can be written as follows, 
\begin{eqnarray*}
     V_n^{\lambda}(\bm{\omega}_n) = 
    \max_{a_n \in \{0,1\}} \left\{
    R(\bm{\omega}_n,a_n) - \lambda a_n 
      \right. \\ \left. 
    + \beta \sum_{o_n \in \mathcal{O}}
    V^{\lambda}(\tau(\bm{\omega}_n,o_n,a_n)) \xi(o_n~| \bm{\omega}_n,a_n)
    \right\}
\end{eqnarray*}
We define Q-belief action value fuction, 
\begin{eqnarray*}
    Q^{\lambda}_n(\bm{\omega}_n,a_n) =  R(\bm{\omega}_n,a_n) - \lambda a  + \beta \sum_{o \in \mathcal{O}}
    V_n^{\lambda}(\tau(\bm{\omega}_n,o_n,a_n)) 
    \\
    \times \xi(o_n~| \bm{\omega}_n,a_n)
\end{eqnarray*}
and 
\begin{eqnarray*}
    V_n^{\lambda}(\bm{\omega}_n) = 
    \max_{a_n \in \{0,1\}}   Q^{\lambda}_n(\bm{\omega}_n,a_n). 
\end{eqnarray*}
Then, the set $U_0(\lambda)$ is defined by 
\begin{eqnarray*}
U_0(\lambda) := \left\{\bm{\omega}_n \in  \Delta ~|~  Q_n^{\lambda}(\bm{\omega_n},a_n=1) \leq  Q_n^{\lambda}(\bm{\omega}_n,a_n=0) \right\} 
\end{eqnarray*}
Next we define the indexability using this set. 
\begin{definition}[Indexability \cite{Whittle88}]
	%Indexability for each arm is defined as set of states at which not playing is optimal choice and this set  monotonically increases with subsidy $W$, that is, 
	As subsidy $\lambda$ increases from $-\infty$ to $+\infty,$  $U_{0}(W)$ increases from $\emptyset$ to full set $\Delta.$ 
	\label{def:indexability-single-dim}	
\end{definition} 
To show indexability we require that whenever $\lambda_2 > \lambda_1$, it implies $U_0(\lambda_1) \subseteq U_0(\lambda_2).$ 
Without structural assumptions, it is non-trivial to show indexability. Often this is done by proving a threshold type optimal policy. When the state is not perfectly observable for all actions, optimal threshold policies are difficult to show, and so is indexability. 

We now define the Whittle index.
%  and present Monte-Carlo rollout policy algorithm. 
\begin{definition}[Whittle index \cite{Whittle88}]
	If an arm $n$ is indexable and is in state $\bm{\omega}_n \in \Delta,$ then its Whittle
	index, $\tilde{\lambda}(\bm{\omega}_n),$ is
	\begin{eqnarray*}
	\tilde{\lambda}(\bm{\omega}_n) := \inf_{\lambda}\{\lambda:  Q_n^{\lambda}(\bm{\omega}_n,a_n = 1) - Q_n^{\lambda}(\bm{\omega}_n,a_n =0) = 0 \} .
	\end{eqnarray*}
	\label{def:whittleind} 
\end{definition}
If an arm satisfies the indexability condition, then it is called as indexable arm. Using Whittle index one is required to compute the index for a given belief state $\bm{\omega}_n$ for $n$th arm. This has to be done for all arms. Arms with highest index under budget constraints are played at each time instant. 

Index computation is non-trivial even after showing indexability. Due to the belief simplex and partial observability, value function computation is hard for POMDPs. Most often, in these models, explicit closed form expressions are difficult, except in special cases where the state is perfectly observable for one of actions. In some cases, structural properties are exploited to come up with index computation algorithm, \cite{Meshram2021indexability,Akbarzadeh2022indexability}. 
\subsection{Multi-action ($J >2$) PO-RMAB}
Multi action ($\geq 3$) and multi state PO-RMAB is challenging problem. These challenges come from multi-actions, and partial observabilty of the model with no perfect state information.
From Lemma~\ref{lemma:val-fun-decouple}, the dynamic program  for arm $i$  is 
\begin{eqnarray*}
     V_n^{\lambda}(\bm{\omega_n}) = 
    \max_{a_n \in \{0,1,2, \cdots,J-1\}} \left\{
    R(\bm{\omega}_n,a_n) - \lambda a_n 
      \right. \\ \left. 
    + \beta \sum_{o_n \in \mathcal{O}_n}
    V_{n}^{\lambda}(\tau(\bm{\omega}_n,o_n,a_n)) \prob{o_n~| \bm{\omega}_n,a_n}
    \right\}
\end{eqnarray*}
Define
\begin{eqnarray*}
    Q^{\lambda}_n(\bm{\omega}_n,a_n) =  R(\bm{\omega}_n,a_n) - \lambda a_n + \beta \sum_{o \in \mathcal{O}}
    V_n^{\lambda}(\tau(\bm{\omega}_n,o_n,a_n)) 
    \\
    \times \xi(o_n~| \bm{\omega}_n,a_n)
\end{eqnarray*}
and 
\begin{eqnarray*}
     V_n^{\lambda}(\bm{\omega_n}) = 
    \max_{a_n \in \{0,1,2, \cdots,J-1\}}  Q^{\lambda}_n(\bm{\omega}_n,a_n),
\end{eqnarray*}
\begin{eqnarray*}
     \widehat{a}_n^{\lambda}(\bm{\omega_n}) = 
   \arg\max_{a_n \in \{0,1,2, \cdots,J-1\}}  Q^{\lambda}_n(\bm{\omega}_n,a_n) 
\end{eqnarray*}
Then, the set $U_0(\lambda)$ is defined by 
\begin{eqnarray*}
U_n(\lambda,a_n) := \left\{\bm{\omega}_n \in  \Delta ~|~   \widehat{a}_n^{\lambda}(\bm{\omega_n}) \leq a_n \right\} .
\end{eqnarray*}
 It is the collection of belief states for which the optimal action is chosen less than or equal to fixed activity level $a_n \in \{0,1,\cdots,J-1\}.$

\begin{definition}[Full Indexability \cite{Glazebrook2011}] 
	An arm $n$ is fully indexable if   $U_n(\lambda,a_n)$ non-decreasing in $\lambda$ for each $a_n \in \{0,1,\cdots,J-1\}.$
	\label{def:indexability-full}	
\end{definition} 
If all arms are fully indexable, then PO-RMAB is called full
indexable.

\begin{definition}[Whittle Index for multi-action \cite{Glazebrook2011}]
	The Whittle index of  fully indexable  arm $n$ with belief state $\omega_n \in \Delta$ is defined as follows.
  \begin{eqnarray*}
	\widetilde{\lambda}_n(\bm{\omega}_n,a_n) := \inf_{\lambda}\{\lambda:   \bm{\omega}_n \in U_n(\lambda,a_n) \} .
	\end{eqnarray*}
	\label{def:whittleind-full} 
    The index $\widetilde{\lambda}_n(\bm{\omega}_n,a_n)$ depends on activity level $a_n.$
\end{definition} 
\begin{lemma}
    If arm $n$ is fully indexable, then $\widetilde{\lambda}(\bm{\omega}_n,a_n)$ is decreasing in $a_n$ for fixed belief state $\bm{\omega}_n.$
    \label{lemma:full-indexable}
\end{lemma}
Proof is given in Appendix~\ref{proof:lemma:full-indexable}.

For partially observable-RMAB, it is difficult to prove full indexability as it require computation of the value function and showing monotonicity of $U_n(\lambda,a_n)$ in $\lambda$ by fixing activity level $a_n.$  

Now, we have a discussion on POMDP, which is useful in understanding full indexability for multi-action PO-RMAB.
\subsection{Discussion: Structural Results on POMDP}
To prove full indexability, one condition is monotonicity of value function and  threshold type policies for multi-actions. In other words, the difference of the action value functions must be monotone (isotone) in actions. Note that the monotonicity of value functions for POMDP needs stronger structural assumptions as studied in \cite{Lovejoy87,White79, White80}. The algorithms and bounds for POMDPs are discussed in \cite{Lovejoy87,Lovejoy91a,Lovejoy91b}.

Hence, proving indexability is non-trivial and the computation of index is also difficult as there is no closed form expression of the value function. However, under structural assumptions on POMDPs, it can be possible to have simplified expressions in case of three action models. Some work on structural results for POMDPs where models are motivated from machine replacement problems, can be found in \cite{Ross71,Lovejoy87b,Sernik1991,Sernik1991b}. 
Here, it is possible to have full indexability and an index formula. 

In general, our earlier study of heuristic policies is better suited for our problem. In addition to this, one can study Monte-carlo tree search algorithm for multi-action PO-RMAB.

\section{Concluding Remarks and Discussion}
\label{sec:concluding-remark}
We have studied multi-action PO-RMAB using Lagrangian relaxation methods. We presented the Lagrangian bound and a computational approach for Lagrangian bound. We studied properties of value functions and studied a two-timescale stochastic approximation algorithm for Lagrangian bound computation. We also discussed PBVI and rollout policy algorithm. We studied Lagrangian based heuristic policies and greedy policy. Further, we provided a discussion and some insight into indexability conditions for PO-RMAB.  

It is the first step towards solving multi-action PO-RMAB. One can also study the PO-RMAB with a multi-agent framework. There are various potential directions for future work such as the study of  efficient algorithms for Lagrangian bound computation, and also the study of Q-learning algorithm for PO-RMAB. 
We further plan to study Monte Carlo Tree Search for PO-RMAB and Column Generation Approach with LP formulation for PO-RMAB, which have been studied for POMDP, \cite{Walraven2018}. 
Another direction for future work is to explore these models for different applications in recommendation systems, communication systems and robotics.  

\section{Acknowledgment}
%R. Meshram is with the Department of Electrical Engineering, Indian Institute of Technology Madras, Chennai, India. K. Kaza is with the Department of Electrical Engineering and Computer Science, University of Ottawa, Canada.
%
The work of Rahul Meshram is supported from IITM NFIG Grant and SERB grant Project No EEQ/2021/000812. 

\bibliographystyle{IEEE}

\bibliography{restless-bandits}

\appendix

%\subsection{title}
%\rahul{Need to clean  the following: }
\subsection{Derivation of Belief Update Rule}
Detailed derivation is provided for clarity sake. 
We note that from Section~\ref{sec:belief-update-rule}, $\omega^s_n(t)$ is the belief about the state $s$ at time $t$ for $n$th arm, we have following. 
\begin{eqnarray*}
\omega_{n}^s(t) &=& \prob{s_{n}(t) = s~|~H(t), \bm{\omega}_n(0)}. 
\end{eqnarray*}

Given observation  from $n$th arm $o_n(t) = k$ and action of that arm is $a_n(t) = a $ and previous belief state $\bm{\omega}_n(t),$ the belief update for state $s_n(t+1) =s$  at time $t+1$ is given as follows. 
\begin{align*}
   \omega_{n}^s(t+1) = \prob{ s_n(t+1) = s~|~ \bm{\omega}_n(t), a_n(t) = a, o_n(t) = k}.
\end{align*}
We have $\prob{s_n(t+1) = s~|~s_n(t) = s^{\prime},a_n(t)= a_n} = p_{s^{\prime},s}^{a_n}.$ and $\omega_n^s(t)$ is probability being in state $s$ for arm $n.$ The observations is from the state $s_n(t) = s^{\prime}$ in our model. Next using Bayes Rule, we obtain 
\begin{eqnarray*}
     \omega_{n}^s(t+1) =   \prob{ s_n(t+1) = s~|~ \bm{\omega}_n(t), a_n(t) = a_n, o_n(t) = k} \\
     =  \frac{\prob{ s_n(t+1) = s, o_n(t) = k~|~ \bm{\omega}_n(t), a_n(t) = a_n}}{\prob{o_n(t) = k~|~\bm{\omega}_n(t), a_n(t) = a_n}}
\end{eqnarray*}
First, we discuss numerator term:
{\small{
\begin{eqnarray*}
& \prob{ s_n(t+1) = s, o_n(t) = k~|~ \bm{\omega}_n(t), a_n(t) = a_n} = \\
& \sum_{s^{\prime} \in \mathcal{S}}\prob{ s_n(t+1) = s, o_n(t) = k~|~ s_n(t) = s^{\prime}, a_n(t) = a_n}\omega_n^{s^{\prime}}(t)
\end{eqnarray*}
}}
Further,
\begin{eqnarray*}
& \prob{ s_n(t+1) = s, o_n(t) = k~|~ s_n(t) = s^{\prime}, a_n(t) = a_n} = \\
& \prob{o_n(t)=k~|~s_n(t)=s,  a_n(t)=a_n} \times \\
&  \prob{s_n(t+1)=s~|~ s_n(t)= s^{\prime}, a_n(t)=a_n}.
\end{eqnarray*}
We can have 
\begin{eqnarray*}
& \prob{ s_n(t+1) = s, o_n(t) = k~|~ s_n(t) = s^{\prime}, a_n(t) = a_n}  = \\
&  \rho_{k,n}^{s^{\prime},a_n} p_{s^{\prime},s}^{a_n}
\end{eqnarray*}
Hence numerator is 
\begin{eqnarray*}
&    \prob{ s_n(t+1) = s, o_n(t) = k~|~ \bm{\omega}_n(t), a_n(t) = a_n}  = \\
& \sum_{s^{\prime} \in \mathcal{S}}  \rho_{k,n}^{s^{\prime},a_n} p_{s^{\prime},s}^{a_n} \omega_n^{s^{\prime}}(t)
\end{eqnarray*}

Next we consider denominator term: 
\begin{eqnarray*}
 &   \prob{o_n(t) = k~|~\omega_n(t), a_n(t) = a_n} = \\
 &   \sum_{s^{\prime} \in \mathcal{S}}  \prob{o_n(t) = k~|~s_n(t) = s^{\prime}, a_n(t) = a_n} \omega^{s^{\prime}}_n(t)
\end{eqnarray*}
Further, we can get 
\begin{eqnarray*}
 &   \prob{o_n(t) = k~|~\omega_n(t), a_n(t) = a_n} = \\
 &   \sum_{s^{\prime} \in \mathcal{S}}  \rho_{k,n}^{s^{\prime},a_n} \omega^{s^{\prime}}_n(t)
\end{eqnarray*}
Combining numerator and denominator term, we have 
\begin{eqnarray*}
    \omega_{n}^s(t+1)  = \frac{\sum_{s^{\prime} \in \mathcal{S}}  \rho_{k,n}^{s^{\prime},a_n} p_{s^{\prime},s}^{a_n} \omega_n^{s^{\prime}}(t)}{\sum_{s^{\prime} \in \mathcal{S}}  \rho_{k,n}^{s^{\prime},a_n} \omega^{s^{\prime}}_n(t)}. 
\end{eqnarray*}
and 
\begin{eqnarray*}
    \bm{\omega}_n(t+1) = \left[ \omega_{n}^0(t+1), \cdots,  \omega_{n}^{M-1}(t+1) \right].
\end{eqnarray*}
This completes the derivation. 
\qed 
%\begin{align*}
% \omega_{n}^s(t+1)   = \frac{\sum_{s^{\prime}\in\mathcal{S}}\rho^{s^{\prime},a_n}_{k,n} \omega_{n}^{s^{\prime}}(t)p^a_n(s^{\prime},s)}{\sum_{s^{\prime}\in\mathcal{S}} \omega_{n}^{s^{\prime}}(t)\rho^{s^{\prime},a}_{k,n}}.
%\end{align*}
\subsection{Proof of Lemma~\ref{lemma:val-fun-decouple}}
\label{Proof-lemma:val-fun-decouple-appendix}
%\rahul{Need to correct the following}

Denote the expressions on the right hand side of \eqref{eqn:DP-1} and  \eqref{eqn:DP-2} as $\mathcal{E}_1$ and $\mathcal{E}_2,$  respectively. We need to show that 
%substituting $\mathcal{E}_2$ in \eqref{eqn:crmab-Lag_relaxed} gives \eqref{eqn:lemma-Lagrangian_decouple}, i.e. 
$\mathcal{E}_1(\mathcal{E}_2) = \mathcal{E}_2.$ That means, it suffices to show that the following expression $\mathcal{E}_1(\mathcal{E}_2)- \mathcal{E}_2 = 0.$ 
Hence, we want to show that the following expression is equal to 0.
{\small{
\begin{dmath*}
\max\limits_{\bm{a}\in \mathcal{A}} \bigg\lbrace\sum\limits_{n=1}^{N} [R_n(\bm{\omega}_n,a_n)-\lambda a_n] + \lambda B + \beta\sum\limits_{\bm{o}\in {\bm{O}}}\prob{\bm{o}|\bm{\omega},\bm{a}}\bigg[\frac{B\lambda}{1-\beta} + \sum\limits_{n=1}^{N}V^{\lambda}(\tau(\bm{\omega}_n,o_n,a_n)) \bigg] \bigg\rbrace - \frac{B\lambda}{1-\beta} - \sum\limits_{n=1}^{N}V_n^{\lambda}(\bm{\omega}_n).
\end{dmath*}
Using $\sum\limits_{\bm{o}\in S_{\bm{o}}} \prob{\bm{o}|\bm{\omega},\bm{a}}=1$ and rearranging the terms, we have \\
\begin{dmath*}
= - \sum\limits_{n=1}^{N}V_n^{\lambda}(\bm{\omega}_n) + \max\limits_{\mathcal{A}} \bigg\lbrace\sum\limits_{n=1}^{N} [R_n(\bm{\omega}_n,a_n)-\lambda a_n] + \beta\sum\limits_{\bm{o}\in {\bm{O}}}\sum\limits_{n=1}^{N} \prob{\bm{o}|\bm{\omega},\bm{a}}  V^{\lambda}(\tau(\bm{\omega}_n,o_n,a_n))  \bigg\rbrace.
\end{dmath*}
Reordering the summations and suitably expanding, we have
\begin{dmath*}
    = - \sum\limits_{n=1}^{N} V_n^{\lambda}(\bm{\omega}_n) + \max\limits_{\bm{a}\in \mathcal{A}} \bigg\lbrace \sum\limits_{n=1}^{N} [R_n(\bm{\omega}_n,a_n)-\lambda a_n] \\ + \beta\sum\limits_{n=1}^{N}\sum\limits_{{o_n}\in O_n}\sum\limits_{\bm{o}_{-n}\in {\bm{O}_{-n}}}\left[ \prob{\bm{o}|\bm{\omega},\bm{a}} \times V^{\lambda}(\tau(\bm{\omega},\bm{o},\bm{a}))	\right]	\bigg\rbrace ,
\end{dmath*} 
where, $\bm{o}_{-n}$ is the observation vector $\bm{o}$ omitting the $n^{th}$ element and $\bm{O}_{-n}\:= \times_{m\neq n} O_m $.
\begin{dmath*}
= - \sum\limits_{n=1}^{N} V_n^{\lambda}(\bm{\omega}_n) + \max\limits_{\bm{a}\in \mathcal{A}_{\bm{y}}}  \bigg\lbrace \sum\limits_{n=1}^{N} [r_n(\bm{\omega}_n,a_n)-\lambda a_n]  +  \beta\sum\limits_{n=1}^{N}\sum\limits_{{o_n}\in O_n} \left[ \prob{{o}_n|\bm{\omega}_n,a_n}\times V_n^{\lambda}(\tau(\bm{\omega}_n,{o}_n,{a}_n)) \right]\bigg\rbrace \\
= { \sum\limits_{n=1}^{N}\bigg( -V_n^{\lambda}(\bm{\omega}_n) + \max\limits_{a_n\in \mathcal{A}}  \bigg\lbrace[r_n(\omega_n,a_n)-\lambda a_n] }\\ + \beta\sum\limits_{{o_n}\in O_n} \left[ \prob{{o}_n|\bm{\omega}_n,a_n} \times  V_n^{\lambda}(\tau(\bm{\omega}_n,{o}_n,{a}_n))\right]\bigg\rbrace\bigg)\\=0.
\end{dmath*} } }

This completes the proof. \qed 

\subsection{Proof of Proposition~\ref{Prop:Val-fn}}
\label{proof:Prop:Val-fn}

We have the following dynamic program for given belief $\bm{\omega} \in \Delta^N$
\begin{eqnarray}
    V(\bm{\omega}) = \max_{\bm{a} \in \mathcal{A} } \left[ \sum_{n=1}^{N} R(\bm{\omega}_n,a_{n}) + \beta 
\sum_{\bm{o} \in \mathcal{O}} V(\tau(\bm{\omega},\bm{o},\bm{a}) )
\nonumber \right. \\ \left.
\prob{ \bm{o} ~|~\bm{\omega},\bm{a}}
    \right]
%\label{Eqn:opt-dynamic-prog-A}    
\end{eqnarray}
Here, the feasible action set is 
\begin{align*}
    \mathcal{A} = \{ \boldsymbol{a}(t) = (a_{n}(t))_{n=1:N} : a_{n}(t) \in    \{0,1,\dots,J\},
    \\ 
    \sum_{n=1}^{N} a_{n}(t)  \leq B \}.
\end{align*}
Let $\overline{\mathcal{A}} =\{ \bm{a}~|~ \bm{a} \in  \{0,1,2,\cdots,J-1\}^N\}. $

From feasibility of constraints, $\left(B - \sum_{n=1}^{N}a_n\right) \geq 0$ for all $\bm{a} \in \overline{\mathcal{A}} $
After Lagrangian relaxation of the preceding dynamic program in RHS, we have
{\small{
\begin{eqnarray*}
    V(\bm{\omega}) \leq \max_{\bm{a} \in \overline{\mathcal{A}}} \left\{ \sum_{n=1}^{N} R(\bm{\omega}_n,a_n) + \lambda\left(B - \sum_{n=1}^{N}a_n\right)
    \right. \nonumber \\ \left. 
    + \beta \sum_{\bm{o} \in \mathcal{O}} V(\tau(\bm{\omega},\bm{o},\bm{a})) \prob{ \bm{o} ~|~\bm{\omega},\bm{a}}
    \right\} \nonumber \\ 
    %\label{eqn:DP-1-A}
\end{eqnarray*}
}}
Further, $\overline{\mathcal{A}} \subseteq \mathcal{A}.$ Hence 
\begin{eqnarray*}
    V(\bm{\omega}) \leq \max_{\bm{a} \in \mathcal{A}} \left\{ \sum_{n=1}^{N} R(\bm{\omega}_n,a_n) + \lambda\left(B - \sum_{n=1}^{N}a_n\right)
    \right. \nonumber \\ \left. 
    + \beta \sum_{\bm{o} \in \mathcal{O}} V(\tau(\bm{\omega},\bm{o},\bm{a})) \prob{ \bm{o} ~|~\bm{\omega},\bm{a}}
    \right\} \nonumber \\ 
    %\label{eqn:DP-1-A}
\end{eqnarray*}
for $\bm{\omega} \in \Delta^N.$
Let $\mathcal{T}^{\lambda}$ be the Bellman operator, it is given as follows.
\begin{eqnarray*}
  \mathcal{T}^{\lambda}  V(\bm{\omega}) = \max_{\bm{a} \in \mathcal{A}} \left\{ \sum_{n=1}^{N} R(\bm{\omega}_n,a_n) + \lambda\left(B - \sum_{n=1}^{N}a_n\right)
    \right. \nonumber \\ \left. 
    + \beta \sum_{\bm{o} \in \mathcal{O}} V(\tau(\bm{\omega},\bm{o},\bm{a})) \prob{ \bm{o} ~|~\bm{\omega},\bm{a}}
    \right\}
\end{eqnarray*}
We have 
\begin{eqnarray*}
    V(\bm{\omega}) \leq \mathcal{T}^{\lambda}  V(\bm{\omega}).
\end{eqnarray*}
From monotonicity of Bellman operator, we can have 
\begin{eqnarray*}
    V(\bm{\omega}) \leq V^{\lambda}(\bm{\omega}).
\end{eqnarray*}
Hence
\begin{eqnarray*}
    V(\bm{\omega}) \leq \min_{\lambda \geq 0} V^{\lambda}(\bm{\omega}).
\end{eqnarray*}
and 
\begin{eqnarray*}
V(\bm{\omega}) \leq V^{\lambda^*}(\bm{\omega}).
\end{eqnarray*}
\qed 

\subsection{Proof of Lemma \ref{lemma:Vproperties}}
%\rahul{need to correct the proof}

We first provide background Lemma from \cite{Astrom69}.
\begin{lemma}
    If \( f: \mathbb{R}^n_+ \to \mathbb{R}_+ \) is a convex function, then for all \( x \in \mathbb{R}^n_+ \), the function
\[
g(x) = \|x\|_1 f\left( \frac{x}{\|x\|_1} \right)
\]
is also convex.
\label{lemma:convex}
\end{lemma}

Note that we want to show that $V_n^{\lambda}(\bm{\omega}_n)$ is convex in $\bm{\omega}_n$
%Here, for notational simplicity, we remove dependence of value function $V_n$ on $n,$ and simply represent with $V.$  

\subsubsection{Convexity of value function in $\bm{\omega}$}
We now show that the value function $V_n^{\lambda}(\omega)$ is piecewise linear and convex in $\omega.$

The proof is using  mathematical induction method. It is along the lines of \cite{Meshram18}. 
We now denote $V_n^{\lambda}(\bm{\omega}_n)$ with time index $V^{\lambda}_{n,t}(\bm{\omega}_n)$ at time step $t.$ Further, assumed that $\lambda$ is given and fixed. 
\begin{itemize}
    \item Let 
    \begin{eqnarray*}
    V_{n,1}^{\lambda}(\bm{\omega}_n) = \max_{a \in \{0, \ldots, J-1\}} \left\{ R(\bm{\omega}_n, a) - \lambda a \right\} \end{eqnarray*}
 \( V_{n,1}^{\lambda}(\bm{\omega}_n) \) is the maximum of linear functions 
    (since \( R(\bm{\omega}_n, a) \) is linear in \(\bm{\omega}_n \))
    Hence, \( V_{n,1}^{\lambda}(\bm{\omega}_n) \) is piecewise linear and convex.
    
    \item We consider induction hypothesis that $V_{n,t}^{\lambda}(\bm{\omega}_n)$ is piecewise linear and convex. Next show that $V_{n,t+1}^{\lambda}(\bm{\omega}_n)$ is piecewise linear and convex. 
    We can rewrite $ V_{n,t+1}^{\lambda}(\bm{\omega})$ in the following form. 
    \begin{eqnarray*}
      V_{n,t+1}^{\lambda}(\bm{\omega}_n) = 
      \max_{a \in \{0,2,\cdots,J-1\}} 
         \left\{ R(\bm{\omega}_n, a) - \lambda a  +  
         \right. \\ \left. 
         \beta \sum_{k \in \mathcal{O}} 
         V_{n.t}^{\lambda}\left( \frac{\xi_k}{|| \xi_k||_1} \right)  || \xi_k||_1
         \right\}
    \end{eqnarray*}
    Here, we define 
    \begin{eqnarray*}
        \xi_k^i := \sum_{s\in\mathcal{S}}\rho^{s,a}_{k,n}\omega^s p^{a,n}_{s^{\prime},s} 
    \end{eqnarray*}
    \begin{eqnarray*}
        \xi_k = [\xi_k^{1}, \cdots, \xi_k^J]^T.
    \end{eqnarray*}
    \begin{eqnarray*}
        ||\xi_k||_1 = \sum_{s\in\mathcal{S}} \omega^s\rho^{s,a}_{k} 
    \end{eqnarray*}
    Using earlier Lemma~\ref{lemma:convex}, $ V_{n,t+1}^{\lambda}(\bm{\omega}_n)$ is piecewise linear and convex in $\bm{\omega}_n.$ 
    \item By induction,  $V_{n,t}^{\lambda}(\bm{\omega})$ is piecewise linear and convex in $\bm{\omega}_n$ for all $t \geq 1.$ From \cite[Chapter $7$]{BertsekasV195}, we can have $V_{n,t}^{\lambda}(\bm{\omega}_n) \rightarrow V_n^{\lambda}(\bm{\omega}_n) $ as $t \rightarrow \infty$ and $V_n^{\lambda}(\bm{\omega}_n)$ is piecewise linear and convex in $\bm{\omega}_n.$ 

\end{itemize}

\subsubsection{Convexity of value function in $\lambda$}
We here show that $V_n^{\lambda}(\bm{\omega})$ is piecewise convex decreasing in $\lambda.$
Proof is again via induction method, and it is along lines of earlier proof. $V_n^{\lambda}(\bm{\omega})$ is a max of linear functions in $\lambda.$ Hence it is also piecewise linear convex in $\lambda.$ It is also decreasing in $\lambda$ as $\lambda a_j \geq 0,$ for $\lambda \geq 0$ and $a_j =0,\cdots,J-1. $ As $\lambda$ increases to $\infty$, the optimal action is not to play any activity. i.e., $a= 0$ for all time and the optimal reward under this policy is 
\begin{eqnarray*}
    V_n^{\lambda}(\bm{\omega}_n) = \mathbb{E}\left[ \sum_{t=1}^{\infty} \beta^{t-1}R(\bm{\omega}_{n,t}, a_t =0)~|\bm{\omega}_{n,1} = \bm{\omega}_n \right]
\end{eqnarray*}
There is no dependence on $\lambda$ in immediate reward, as $a_j =0,$ for all times. Thus $\lambda \rightarrow \infty$ we can have $\frac{\partial V_n^{\lambda}(\bm{\omega}_n)}{\partial \lambda} \rightarrow 0.$ 

\subsubsection{Lipschitz Property} 
Proof is along the lines of \cite[Theorem~5.1]{Saldi2017}. We describe the proof for partially observable MDP.
Note that $\Delta$ is the set of belief state space. 
\begin{eqnarray*}
    \Delta = \{ \bm{\omega}_n ~|~ \sum_{s=1}^{J}\omega^s_n = 1, 0 \leq \omega^s_n \leq 1 \}
\end{eqnarray*}
Hence $\Delta$ is a simplex of dimension $M-1.$ \\
Define the operator $\mathcal{T}$ is Bellman operator on $B(\Delta),$ 
\begin{eqnarray*}
    \mathcal{T}u(\bm{\omega}_n) =  \max_{a \in \{0,1,2,\cdots,J-1\}} 
         \left\{ R(\bm{\omega}_n, a) - \lambda a  +  
         \right. \\ \left. 
         \beta \sum_{o_n \in \mathcal{O}} 
         u(\tau(\bm{\omega}_n,o_n, a_n)) \xi(\bm{\omega}_n,o_n,a_n)
         \right\}
\end{eqnarray*}
%$T$ is called the Bellman optimality operator. 
Further, $\mathcal{T}$ is a contraction operator and $\mathcal{T}:C_b(\Delta) \rightarrow C_b(\Delta).$ $\mathcal{T}u \in C_b(\Delta).$ Note that $B(\Delta)$ is set of all bounded measurable functions and $C_b(\Delta)$ is set of all continuous real valued functions. Also, 
\begin{eqnarray*}
    || \mathcal{T}u - \mathcal{T}v|| \leq \beta || u -v|| & \mbox{ for all $u,v \in C_b(\Delta).$}
\end{eqnarray*}

Let 
\begin{eqnarray*}
    \tilde{U}(\bm{\omega}_n) :=  \sum_{o_n \in \mathcal{O}} 
         u(\tau(\bm{\omega}_n,o_n, a_n)) \xi(\bm{\omega}_n,o_n,a_n)
\end{eqnarray*}

Let $\bm{\omega}_{n,1}$  and $\bm{\omega}_{n,2}$ are two belief state vectors and $\bm{\omega}_{n,1}, \bm{\omega}_{n,2} \in \Delta,$ next we want to obtain bound on the following. 
\begin{eqnarray*}
       \bigg \vert \tilde{U}(\bm{\omega}_{n,1}) - \tilde{U}(\bm{\omega}_{n,2}) \bigg \vert \leq K L_2 d_{\Delta}( \bm{\omega}_{n,1}, \bm{\omega}_{n,2}).
\end{eqnarray*}

Secondly, we bound the following
\begin{eqnarray*}
     \bigg\vert \mathcal{T} u(\bm{\omega}_{n,1}) - \mathcal{T} u(\bm{\omega}_{n,2}) \bigg\vert \leq  (L_1  + \beta K L_2) d_{\Delta}(\bm{\omega}_1, \bm{\omega}_2)
\end{eqnarray*}

Note that $\mathcal{T}$ is contraction operator  and $\mathcal{T}u \in \mathrm{Lip}(\Delta,L_1+ KL_2).$
By recursion, we obtained $\mathcal{T}^tu = \mathcal{T}(\mathcal{T}^{t-1}u)$ and it converges to the value function $V$ by the Banach fixed point theorem.  Hence by induction method, we can have for $t\geq 1$
\begin{eqnarray*}
    \mathcal{T}^tu \in \mathrm{Lip}(\Delta,\widetilde{L}_t)
\end{eqnarray*}
Here,
\begin{eqnarray*}
    \widetilde{L}_t = L_1 + \sum_{i=1}^{t-1} (\beta L_2)^2 + K (\beta L_2)^t
\end{eqnarray*}
If we choose $K < L_1$ then $ \widetilde{L}_t \leq  \widetilde{L}_{t+1}$ for all $t$ and therefore $ \widetilde{L}_t \rightarrow \frac{L_1}{1- \beta L_2}$ since $L_2 \beta < 1.$
Therefore the value function $V \in \mathrm{Lip}(\Delta,\frac{L_1}{1-\beta L_2}).$ Here $\mathrm{Lip}(\Delta,\frac{L_1}{1-\beta L_2}).$ is closed with respect to sup norm $||\cdot||.$
Hence the value function is Lipschitz with constant $\frac{L_1}{1-\beta L_2}.$ 

This completes the proof. \qed. 

We now provide proof for intermediate steps. 
\begin{proposition}
\begin{eqnarray*}
    \bigg \vert \tilde{U}(\bm{\omega}_{n,1}) - \tilde{U}(\bm{\omega}_{n,2}) \bigg \vert \leq K L_2 d_{\Delta}( \bm{\omega}_{n,1}, \bm{\omega}_{n,2}).
\end{eqnarray*}
\end{proposition}
Proof is as follows. 
\begin{eqnarray*}
   & \bigg \vert \tilde{U}(\bm{\omega}_{n,1}) - \tilde{U}(\bm{\omega}_{n,2}) \bigg \vert  = \\
 &   \bigg \vert \sum_{o_n \in \mathcal{O}} \left(  u(\tau(\bm{\omega}_{n,1},o_n, a_n)) \xi(\bm{\omega}_{n,1},o_n,a_n) -
 \right.\\ &  \left. 
  u(\tau(\bm{\omega}_{n,2},o_n, a_n)) \xi(\bm{\omega}_{n,2},o_n,a_n)\right)  \bigg \vert .
\end{eqnarray*}
There $K$ number of observations, we can upper bound equality as follows.
\begin{eqnarray*}
   & \bigg \vert \tilde{U}(\bm{\omega}_{n,1}) - \tilde{U}(\bm{\omega}_{n,2}) \bigg \vert  = \\
 &  K \max_{o_n \in \mathcal{O}} \bigg \vert \left(  u(\tau(\bm{\omega}_{n,1},o_n, a_n)) \xi(\bm{\omega}_{n,1},o_n,a_n) -
 \right.\\ &  \left. 
  u(\tau(\bm{\omega}_{n,2},o_n, a_n)) \xi(\bm{\omega}_{n,2},o_n,a_n)\right)  \bigg \vert .
\end{eqnarray*}
Note that $\xi(\bm{\omega}_{n,1},o_n,a_n)$ and $ \xi(\bm{\omega}_{n,2},o_n,a_n)$ are probabilities and less than $1,$ and hence 
\begin{eqnarray*}
   & \bigg \vert \tilde{U}(\bm{\omega}_{n,1}) - \tilde{U}(\bm{\omega}_{n,2}) \bigg \vert  \leq  \\
 &  K \max_{o_n \in \mathcal{O}} \bigg \vert   u(\tau(\bm{\omega}_{n,1},o_n, a_n)) -   u(\tau(\bm{\omega}_{n,2},o_n, a_n))  \bigg \vert \times 
\\ 
&   d_{\Delta}( \xi(\bm{\omega}_{n,1},o_n,a_n), \xi(\bm{\omega}_{n,2},o_n,a_n) )
\end{eqnarray*}
Because $d_{\Delta} \leq  1,$  we can have  
\begin{eqnarray*}
   & \bigg \vert \tilde{U}(\bm{\omega}_{n,1}) - \tilde{U}(\bm{\omega}_{n,2}) \bigg \vert  \leq  \\
 &  K \max_{o_n \in \mathcal{O}} \bigg \vert   u(\tau(\bm{\omega}_{n,1},o_n, a_n)) -   u(\tau(\bm{\omega}_{n,2},o_n, a_n))  \bigg \vert
\end{eqnarray*}
Next using $u$ be Lipchitz with parameter $L$ we can get 
\begin{eqnarray*}
   & \bigg \vert \tilde{U}(\bm{\omega}_{n,1}) - \tilde{U}(\bm{\omega}_{n,2}) \bigg \vert  \leq  \\
 &  K L  \max_{o_n \in \mathcal{O}} \bigg \vert \tau(\bm{\omega}_{n,1},o_n, a_n)- \tau(\bm{\omega}_{n,2},o_n, a_n) \bigg \vert .
\end{eqnarray*}
Next we can have following.
\begin{eqnarray*}
   \bigg \vert \tilde{U}(\bm{\omega}_{n,1}) - \tilde{U}(\bm{\omega}_{n,2}) \bigg \vert  \leq  
  K L_2 d_{\Delta}(\bm{\omega}_{n,1}, \bm{\omega}_{n,2})  .
\end{eqnarray*}
\qed 

\begin{proposition}
    \begin{eqnarray*}
        \bigg\vert \mathcal{T} u(\bm{\omega}_{n,1}) - \mathcal{T} u(\bm{\omega}_{n,2}) \bigg\vert \leq  (L_1  + \beta K L_2) d_{\Delta}(\bm{\omega}_1, \bm{\omega}_2)
    \end{eqnarray*}
\end{proposition} 
%\begin{ieeeproof}
Proof is as follows. 
%\rahul{Need to clean the following.}
\begin{eqnarray*}
     \bigg\vert \mathcal{T} u(\bm{\omega}_{n,1}) - \mathcal{T} u(\bm{\omega}_{n,2}) \bigg\vert \leq \max_{a} \left\{ 
   \vert R(\bm{\omega}_{n,1},a) - R(\bm{\omega}_{n,2},a) \vert 
   \right. \\  \left. 
   + \beta   \bigg \vert \tilde{U}(\bm{\omega}_{n,1}) - \tilde{U}(\bm{\omega}_{n,2}) \bigg \vert
   \right\}. 
\end{eqnarray*}
Further, we can have 
\begin{eqnarray*}
   \bigg\vert \mathcal{T} u(\bm{\omega}_{n,1}) - \mathcal{T} u(\bm{\omega}_{n,2}) \bigg\vert   \leq  L_1 d_{\Delta}(\bm{\omega}_{n,1}, \bm{\omega}_{n,2}) + \\ \beta K L_2d_{\Delta}(\bm{\omega}_{n,1}, \bm{\omega}_{n,2}).
\end{eqnarray*}
Hence 
\begin{eqnarray*}
   \bigg\vert \mathcal{T} u(\bm{\omega}_{n,1}) - \mathcal{T} u(\bm{\omega}_{n,2}) \bigg\vert   \leq  (L_1  + \beta K L_2) d_{\Delta}(\bm{\omega}_{n,1}, \bm{\omega}_{n,2}).
\end{eqnarray*}
\qed 
%\end{ieeeproof}

\subsection{Proof of Lemma~\ref{lemma:opt-policy}}
\label{proof:lemma:opt-policy}
Proof is given below. 
\begin{eqnarray*}
    V(\bm{\omega}) = \max_{\bm{a} \in \mathcal{A} } \left[ \sum_{n=1}^{N} R(\bm{\omega}_n,a_{n}) +  
 %   \right. \\ \left. 
    \beta 
\sum_{\bm{o} \in \mathcal{O}} V(\tau(\bm{\omega},\bm{o},\bm{a}) )
\nonumber \right. \\ \left.
\prob{ \bm{o} ~|~\bm{\omega},\bm{a}}
    \right].
%\label{Eqn:opt-dynamic-prog-r}    
\end{eqnarray*}
This can be written as follows. 
\begin{eqnarray*}
    V(\bm{\omega}) = \max_{\bm{a} \in \mathcal{A} } \left[ \sum_{n=1}^{N} R(\bm{\omega}_n,a_{n}) + \beta 
\mathbb{E}\left[ V(\bm{\omega}^{\prime})~|~\bm{\omega},\bm{a} \right] 
%\nonumber \right. \\ \left.
%\prob{ \bm{o} ~|~\bm{\omega},\bm{a}}
    \right].
%\label{Eqn:opt-dynamic-prog-r}    
\end{eqnarray*}
From Proposition~\ref{Prop:Val-fn} and Eqn.~\eqref{eqn:DP-2}, we obtain
\begin{eqnarray*}
 \mathbb{E}\left[ V(\bm{\omega}^{\prime})~|~\bm{\omega},\bm{a} \right]  &\leq & \mathbb{E}\left[ V^{\lambda}(\bm{\omega}^{\prime})~|~\bm{\omega},\bm{a} \right]   \\
 &=& \sum_{n=1}^{N} \mathbb{E}\left[ V_n^{\lambda}(\bm{\omega}^{\prime}_n) ~|~\bm{\omega},\bm{a} \right] + \frac{B \lambda}{1-\beta} . 
\end{eqnarray*}

Note that for given $\lambda,$ preceding equation is nonlinear separable problem over linear  constraint. 
Further, we can obtain 
\begin{eqnarray*}
\bm{a}^{*}(\bm{\omega},\lambda) = \arg \max_{\bm{a} \in \mathcal{A} } \left[ \sum_{n=1}^{N} \left(R(\bm{\omega}_n,a_{n}) + \beta 
\right. \right. \\ \left. \left. 
\sum_{o_n \in \mathcal{O}_n}
    V_{n}^{\lambda}(\tau(\bm{\omega}_n,o_n,a_n)) \xi(\bm{\omega},o_n,a_n)
\right) \right] .
\end{eqnarray*}
Hence for $\bm{\omega}^{\prime}_n= \tau(\bm{\omega}_n,o_n,a_n).$ $n=1,2, \cdots, N$ we have 
\begin{eqnarray*}
\bm{a}^{*}(\bm{\omega},\lambda) = \arg \max_{\bm{a} \in \mathcal{A} } \left[ \sum_{n=1}^{N} \left(R(\bm{\omega}_n,a_{n}) + 
   \right. \right. \\ \left.  \left. 
\beta \mathbb{E}\left[ V_n^{\lambda}(\bm{\omega}^{\prime}_n) ~|~\bm{\omega},\bm{a} \right]  \right) \right] .
\end{eqnarray*}  
\qed 

\subsection{Proof of Lemma~\ref{lemma:full-indexable}}
\label{proof:lemma:full-indexable}
%\subsection{PBVI Algorithm and its analysis}

If arm is fully indexable, then the Whittle index
is minimum amount of subsidy $\lambda$ required such that optimal
actions or activity level less than $a$ for given belief state $\bm{\omega}_n$ and
activity level $a_n.$ It is the subsidy at arm $n$ for raising activity
level $a_n$ to $a_n + 1$  for given belief state $\bm{\omega}_n.$ The subsidy is less than
$\widetilde{\lambda}_n(\bm{\omega}_n, a_n),$ that means reward from low activity level is less.
Hence higher activity levels are preferable. If the subsidy is
higher than index $\widetilde{\lambda}_n(\bm{\omega}_n, a_n),$ then higher activity level are
not preferable. One can define $\widetilde{\lambda}_n(\bm{\omega}_n, a_n = J-1) = 0$  for all
$\bm{\omega}_n \in \Delta.$ Then as activity level increases for fixed $\bm{\omega}_n$ from $a_n$
to $a_n+1$ , and this discussion it is clear that
$\widetilde{\lambda}_n(\bm{\omega}_n, a_n)$ is decreasing in activity level $a_n.$

\qed

\end{document}